\documentclass{article}

\usepackage{arxiv}

\usepackage[utf8]{inputenc} 
\usepackage[T1]{fontenc}    
\usepackage{amsmath,amssymb,amsfonts}%
\usepackage{amsthm}%
\usepackage{algorithm}%
\usepackage{algorithmicx}%
\usepackage{algpseudocode}%
\usepackage{multirow}
\usepackage{hyperref}       
\usepackage{cleveref}
\usepackage{url}            
\usepackage{booktabs}       
\usepackage{amsfonts}       
\usepackage{nicefrac}       
\usepackage{microtype}      
\usepackage{lipsum}		
\usepackage{graphicx}
\usepackage[table]{xcolor}
\usepackage{xcolor}
\usepackage[numbers]{natbib}
\usepackage{doi}

\title{PRKAN: Parameter-Reduced Kolmogorov-Arnold Networks}

\author{ \href{https://orcid.org/0000-0003-0321-5106}{\includegraphics[scale=0.06]{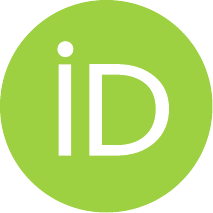}\hspace{1mm}Hoang-Thang Ta}
	\\ Department of Information Technology, Dalat University\\
	Lam Dong, Vietnam \\
	\texttt{thangth@dlu.edu.vn} \\
	   \AND
	\href{https://orcid.org/0000-0002-3991-2037}{\includegraphics[scale=0.06]{orcid.pdf}\hspace{1mm}Duy-Quy Thai} \\
	  Department of Information Technology, Dalat University \\
	Lam Dong, Vietnam \\
	\texttt{quytd@dlu.edu.vn} \\
     \AND
	\href{https://orcid.org/0000-0002-2903-4373}{\includegraphics[scale=0.06]{orcid.pdf}\hspace{1mm}Anh Tran} \\
	  FPT University \\
	Danang, Vietnam \\
       \texttt{anhtn35@fe.edu.vn} \\
	  \texttt{} \\
     \AND
	\href{https://orcid.org/0000-0003-3901-3522}{\includegraphics[scale=0.06]{orcid.pdf}\hspace{1mm}Grigori Sidorov} \\
	  Centro de Investigación en Computación (CIC), Instituto Politécnico Nacional (IPN) \\
	CDMX, Mexico \\
	  \texttt{sidorov@cic.ipn.mx} \\
	 \AND
	\href{https://orcid.org/0000-0001-7845-9039}{\includegraphics[scale=0.06]{orcid.pdf}\hspace{1mm}Alexander Gelbukh} \\
	Centro de Investigación en Computación (CIC), Instituto Politécnico Nacional (IPN) \\
	 CDMX, Mexico \\
	 \texttt{gelbukh@cic.ipn.mx} \\
}

\renewcommand{\shorttitle}{\textit{arXiv} Template}

\hypersetup{
pdftitle={A template for the arxiv style},
pdfsubject={q-bio.NC, q-bio.QM},
pdfauthor={David S.~Hippocampus, Elias D.~Striatum},
pdfkeywords={First keyword, Second keyword, More},
}

\begin{document}
\maketitle

\begin{abstract}

Kolmogorov-Arnold Networks (KANs) represent an innovation in neural network architectures, offering a compelling alternative to Multi-Layer Perceptrons (MLPs) in models such as Convolutional Neural Networks (CNNs), Recurrent Neural Networks (RNNs), and Transformers. By advancing network design, KANs drive groundbreaking research and enable transformative applications across various scientific domains involving neural networks. However, existing KANs often require significantly more parameters in their network layers than MLPs. To address this limitation, this paper introduces PRKANs (\textbf{P}arameter-\textbf{R}educed \textbf{K}olmogorov-\textbf{A}rnold \textbf{N}etworks), which employ several methods to reduce the parameter count in KAN layers, making them comparable to MLP layers. Experimental results on the MNIST and Fashion-MNIST datasets demonstrate that PRKANs outperform several existing KANs, and their variant with attention mechanisms rivals the performance of MLPs, albeit with slightly longer training times. Furthermore, the study highlights the advantages of Gaussian Radial Basis Functions (GRBFs) and layer normalization in KAN designs. The repository for this work is available at: \url{https://github.com/hoangthangta/All-KAN}.

\end{abstract}

\section{Introduction}
MLPs have been one of the key components in modern neural network architectures for years. Their simplicity makes them widely used for capturing complex relationships through multiple layers of non-linear transformations. However, their role has been reconsidered recently with the introduction of Kolmogorov-Arnold Networks (KANs)~\cite{liu2024kan,liu2024kan2.0}. In these papers, fixed activation functions used in MLPs are described as "nodes," and the authors proposed replacing them with learnable activation functions like B-splines, referred to as "edges", to improve performance in mathematical and physical examples. 
To address Hilbert's 13th problem~\cite{sternfeld2006hilbert}, the Kolmogorov–Arnold Representation Theorem (KART)~\cite{kolmogorov1957representation} was introduced. It posits that any continuous function involving multiple variables can be decomposed into a sum of continuous functions of single variables, thus inspiring the creation of KANs.

The work of \citet{liu2024kan} on KANs has inspired numerous studies exploring the use of various basis and polynomial functions as replacements for B-splines~\cite{li2024kolmogorov,athanasios2024,ta2024bsrbf,abueidda2024deepokan,torchkan,ss2024chebyshev,xu2024fourierkan,bozorgasl2024wav,teymoor2024exploring}, investigating the model's performance compared to MLPs. Several studies have shown that KANs do not always outperform MLPs when using the same training parameters~\cite{yu2024kan, zeng2024kanversusmlpirregular}. Moreover, while KANs achieve better performance than MLPs with the same network structure, they often require a significantly larger number of parameters~\cite{ta2024bsrbf, ta2024fc, yang2024activation, moradi2024kolmogorov, sohail2024training}. Conversely, in some cases, KANs use fewer parameters than MLPs but demand significantly more training time~\cite{shuai2024physics}. This suggests that KANs may underperform compared to MLPs in certain tasks when the same network configuration, including both trainable parameters and network structure, is applied. Therefore, to ensure a fair comparison, it is necessary to find methods to reduce the number of parameters in KAN layers while maintaining the same network structure as MLPs. This also helps to clarify the true power of KANs when the same parameters and network structure are compared to MLPs.


In this paper, we propose methods that utilize attention mechanisms, dimension summation, feature weight vectors, and convolutional/pooling layers to effectively reduce the number of parameters in KAN layers. These methods not only preserve the network structure but also enhance feature representation, highlight crucial information, capture spatial dependencies, and, at the same time, minimize parameterization. As a result, each KAN layer corresponds to an MLP layer in terms of structure and parameter count, making KANs more efficient and compact. We gather these methods to introduce a novel network called PRKAN (\textbf{P}arameter-\textbf{R}educed \textbf{K}olmogorov-\textbf{A}rnold \textbf{N}etworks). PRKANs integrated attention mechanisms exhibit competitive performance compared to MLPs on two image datasets, MNIST and Fashion-MNIST, and is anticipated to pave the way for future research on reducing KAN parameters. In summary, our main contributions are:
\begin{itemize}
    \item Develop PRKANs to showcase various methods for reducing the number of parameters in KANs while preserving a network structure similar to MLP layers.
    \item Demonstrate the competitive performance of PRKANs compared to MLPs by training models on two image datasets, MNIST and Fashion-MNIST. Also, explore components that can contribute to the output performance of PRKANs.
\end{itemize}

Aside from this section, the paper is organized as follows: Section 2 discusses related work on KART and KANs. Section 3 details our methodology, covering KART, the design of the KAN architecture, several existing KANs, the parameters used in MLPs and KANs, PRKANs, and data normalization. Section 4 presents our experiments, comparing PRKANs with MLPs and other existing KANs using data from the MNIST and Fashion-MNIST datasets. In addition, this section includes several ablation studies on the activation functions used in PRKANs, a comparison between RBF and B-splines, and a suggestion for the positioning of data normalization in PRKANs. Section 5 discusses some limitations of this paper. Lastly, Section 6 offers our conclusions and potential directions for future research.

\section{Related Works}\label{sec_related_works}

\subsection{KART}
In 1957, Kolmogorov addressed Hilbert’s 13th problem by proving that any multivariate continuous function can be represented as a combination of single-variable functions and sums, a concept known as KART (Kolmogorov–Arnold Representation Theorem)~\cite{kolmogorov1957representation,braun2009constructive}. This idea has been influential in the development of neural networks~\cite{zhou2022treedrnet,leni2013kolmogorov,lai2021kolmogorov,van2022kasam}. However, there is ongoing debate about its practical application in neural network design. \citet{girosi1989representation} raised concerns about KART, suggesting that the inner function $\phi_{q,p}$ in \Cref{eq:kart} may be highly non-smooth~\cite{vitushkin1954hilbert}, potentially affecting the smoothness of the function $f$, which is essential for generalization and noise resistance in neural networks. \citet{lin1993realization} supports this statement based on their discussion, stating that Kolmogorov’s theorem is irrelevant to neural networks for function approximation. On the other hand, \citet{kuurkova1991kolmogorov} argued in favor of KART's relevance to neural networks, showing that linear combinations of affine functions, coupled with certain sigmoidal functions, can effectively approximate all single-variable functions.

\subsection{Development of KAN}

Although KART has a long history of application in neural networks, it did not gain significant attention in the research community until the recent work by \citet{liu2024kan,liu2024kan2.0}. They proposed moving beyond the rigid structure of KART and extending it to create KANs with additional neurons and layers. This approach aligns with our perspective, as it effectively addresses the challenge of non-smooth functions that arise when applying KART in neural networks. As a result, KANs have the potential to surpass MLPs in terms of both accuracy and interpretability, especially for small-scale AI + Science tasks. However, the revival of KANs has been met with criticism from \citet{dhiman2024kan}, who argue that KANs are essentially MLPs that use spline-based activation functions, unlike traditional MLPs which rely on fixed activation functions. 

By offering a fresh perspective on neural network design, KANs have showcased their effectiveness in numerous studies, addressing various problems, including expensive computations~\cite{hao2024first}, differential equations~\cite{wang2024kolmogorov, koenig2024kan}, keyword spotting~\cite{xu2024effective}, mechanics challenges~\cite{abueidda2024deepokan}, quantum computing~\cite{kundu2024kanqas, wakaura2024variational, troy2024sparks}, survival analysis~\cite{knottenbelt2024coxkan}, time series forecasting~\cite{genet2024tkan, xu2024kolmogorov, vaca2024kolmogorov, genet2024temporal, han2024kan4tsf}, and vision tasks~\cite{li2024u, cheon2024demonstrating, ge2024tc}. From that, the role of KAN can be highlighted for its versatility and strength in tackling complex real-world problems across multiple scientific and engineering domains.

Many basis and polynomial functions are used in novel KANs~\cite{somvanshi2024survey}, particularly those adept at handling curves, such as B-splines~\cite{de1972calculating} (Original KAN~\cite{liu2024kan}, EfficientKAN~\cite{Blealtan2024}, BSRBF-KAN~\cite{ta2024bsrbf}), Gaussian Radial Basis Functions (GRBFs) (FastKAN~\cite{li2024kolmogorov}, DeepOKAN~\cite{abueidda2024deepokan}, BSRBF-KAN~\cite{ta2024bsrbf}), Chebyshev polynomials (TorchKAN~\cite{torchkan}, Chebyshev KAN~\cite{ss2024chebyshev}), Legendre polynomials (TorchKAN~\cite{torchkan}), Fourier transforms (FourierKAN\footnote{https://github.com/GistNoesis/FourierKAN/}, FourierKAN-GCF~\cite{xu2024fourierkan}), wavelets~\cite{bozorgasl2024wav,seydi2024unveiling}, rational functions~\cite{aghaei2024rkan}, fractional Jacobi functions~\cite{aghaei2024fkan}, and other polynomial functions~\cite{teymoor2024exploring}. In addition, there are several works that utilize trigonometric functions (LArctan-SKAN~\cite{chen2024larctan}, LSS-SKAN~\cite{chen2024lss}) and activation functions (ReLU-KAN~\cite{qiu2024relu}, Reflection Switch Activation Function (RSWAF) in FasterKAN~\cite{athanasios2024}) design their KAN architectures.

With the ability to replace MLPs, KANs are also integrated into other neural networks such as autoencoders~\cite{moradi2024kolmogorov}, GNNs~\cite{bresson2024kagnns, de2024kolmogorov, zhang2024graphkan}, Reinforcement Learning~\cite{kich2024kolmogorov}, Transformer~\cite{yang2024kolmogorov}, CNNs, and Recurrent Neural Networks (RNNs). When integrated with CNNs, KANs can replace convolutional layers, MLP layers, or both, and even various combinations of them~\cite{abd2024ckan, bodner2024convolutional}, opening up numerous design options. In RNNs, KAN layers do not play the main role; instead, they must be combined with linear weights, fully connected layers, and other components such as biases and previous hidden states to form network structures~\cite{genet2024tkan, genet2024temporal, danish2025kolmogorov}.

\subsection{Parameter Reduction in KANs}

Parameter reduction aims to create models that use fewer resources, both computational and storage, while maintaining or improving performance by eliminating redundant components. This is particularly valuable for large language models, which often come with significant computational and memory costs. KANs have demonstrated efficient parameter usage in tasks such as satellite traffic forecasting~\cite{vaca2024kolmogorov} and quantum architecture search~\cite{kundu2024kanqas}, outperforming MLPs without requiring parameter reduction. However, in various tasks with comparable parameters and computational complexity, MLPs generally outperform KANs, with the exception of symbolic formula representation tasks~\cite{yu2024kan}. In software and hardware implementation, MLP remains an effective approach to achieve accuracy, as KANs do not achieve good results on highly complex datasets while consuming substantially more hardware resources~\cite{le2024exploring}. Although techniques such as tensor decomposition~\cite{ji2019survey, kolda2009tensor}, matrix factorization~\cite{sainath2013low}, and advanced pruning~\cite{cheng2024survey, vadera2022methods} have been developed to reduce parameters, only a few studies specifically address this issue in KANs.

\citet{bodner2024convolutional} introduced Convolutional Kolmogorov-Arnold Networks (Convolutional KANs), integrating learnable non-linear activation functions into convolutions, achieving accuracy similar to CNNs with half the parameters, thus improving learning efficiency. CapsuleKAN enhances precision and parameter efficiency in traditional capsule networks by employing ConvKAN and LinearKAN~\cite{mou2024efficient}. ConvKAN applies B-spline convolutions to improve feature extraction, while LinearKAN uses B-splines as activation functions to capture non-linearities with fewer parameters. Two studies by the same authors introduced Single-Parameterized Kolmogorov-Arnold networks (SKANs), which use basis functions with a single learnable parameter~\cite{chen2024lss, chen2024larctan}. They proposed various variants of SKAN, including LSS-SKAN, LSin-SKAN, LCos-SKAN, and LArctan-SKAN, demonstrating improvements in both accuracy and computational efficiency, with the same approximately 80,000 parameters in all models compared. However, they did not compare SKANs with standalone MLPs; instead, they combined SKANs with other networks, such as MLP+rKAN and MLP+fKAN. Furthermore, in their implementation code, they employed dimension summation to reduce the number of parameters, which may not be the most effective method\footnote{https://github.com/chikkkit/skan\_library}.

We also identified studies that, while not directly focused on parameter reduction in KANs, provide valuable insights related to this topic. For instance, rather than focusing on KANs—which require significantly more parameters and perform less effectively in data-scarce domains—\citet{pourkamali2024kolmogorov} explore MLPs with parameterized activation functions for each neuron. Their experiments show that MLPs achieve higher accuracy than KANs with only a modest increase in parameters. To improve training speed, \citet{qiu2024relu} introduced ReLU-KAN, replacing B-splines with matrix addition, dot products, and ReLU activation to enable efficient GPU parallelization. It also uses exactly two parameters traditionally associated with KANs: grid size and spline order. This method accelerates backpropagation, enhances accuracy, and improves resilience to catastrophic forgetting, all while keeping parameter settings constant. \citet{zinage2024dkl} integrated deep kernel learning (DKL) into KANs (DKL-KAN) and MLPs (DKL-MLP), designing DKL-KAN variants: one with the same neurons and layers as DKL-MLP, and another with a similar number of trainable parameters. They found that DKL-KAN performs better on small datasets, especially in modeling discontinuities and estimating uncertainty, while DKL-MLP excels in scalability and accuracy on larger datasets.

To the best of our knowledge, no existing research compares KANs and MLPs using identical network structures and the same number of trainable parameters. This absence of direct comparison highlights the uniqueness and value of our research, as it enables us to focus on developing effective parameter reduction methods specifically tailored for KANs.

\section{Methodology}
\label{sec:methodology}

\subsection{Kolmogorov-Arnold Representation Theorem}

A KAN is grounded in KART, which states that any continuous multivariate function $f$, defined in a bounded domain, can be expressed as a finite combination of continuous single-variable functions and their summations~\cite{chernov2020gaussian,schmidt2021kolmogorov}. For a set of variables $\mathbf{x} = \{x_1, x_2, \ldots, x_n\}$, where $n$ is the number of variables, the continuous multivariate function $f(\mathbf{x})$ can be represented as:

\begin{equation}
\begin{aligned}
f(\mathbf{x}) = f(x_1, \ldots, x_n) = \sum_{q=1}^{2n+1} \Phi_q \left( \sum_{p=1}^{n} \phi_{q,p}(x_p) \right) 
\end{aligned}
\label{eq:kart}
\end{equation}
This equation involves two types of summations: the outer sum and the inner sum. The outer sum, $\sum_{q=1}^{2n+1}$, combines $2n+1$ terms of $\Phi_q$ ($\mathbb{R} \to \mathbb{R}$), which are continuous functions. The inner sum, on the other hand, aggregates $n$ terms for each $q$, where each term $\phi_{q,p}$ ($\phi_{q,p} \colon [0,1] \to \mathbb{R}$) represents a continuous function of a single variable $x_p$.

\subsection{Design of KANs}
\label{KAN_design}


An MLP is composed of a series of affine transformations followed by non-linear activation functions. Beginning with an input $\mathbf{x}$, the network processes it through multiple layers, each characterized by a weight matrix and a bias vector. For a network with $L$ layers (indexed from $0$ to $L-1$), the transformation at layer $l$ is given by:

\begin{equation}
\begin{aligned}
\text{MLP}(\mathbf{x}) &= (W_{L-1} \circ \sigma \circ W_{L-2} \circ \sigma \circ \cdots \circ W_1 \circ \sigma \circ W_0) \mathbf{x} \\
&= \sigma \left( W_{L-1} \sigma \left( W_{L-2} \sigma \left( \cdots \sigma \left( W_1 \sigma \left( W_0 \mathbf{x} \right) \right) \right) \right) \right)
\end{aligned}
\label{eq:mlp}
\end{equation}


Inspired by KART, \citet{liu2024kan} introduced KANs and advocated for expanding the width and depth of the network to improve its capabilities. This requires identifying appropriate functions $\Phi_q$ and $\phi_{q,p}$ as in \Cref{eq:kart}. A typical KAN with \(L\) layers processes the input \(\mathbf{x}\), where it is sequentially transformed by function matrices \(\Phi_0, \Phi_1, \dots, \Phi_{L-1}\), leading to the output \(\text{KAN}(\mathbf{x})\), as shown in:

\begin{equation}
\begin{aligned}
\text{KAN}(\mathbf{x}) = (\Phi_{L-1} \circ \Phi_{L-2} \circ \cdots \circ \Phi_1 \circ \Phi_0)\mathbf{x}
\end{aligned}
\label{eq:kan}
\end{equation}
which $\Phi_{l}$, called the function matrix, contains a set of pre-activations at the $l^{th}$ KAN layer. Consider the \(i^{th}\) neuron in the \(l^{th}\) layer and the \(j^{th}\) neuron in the \((l+1)^{th}\) layer. The activation function \(\phi_{l,i,j}\) establishes the connection between \((l, i)\) and \((l+1, j)\) as:


\begin{equation}
\begin{aligned}
\phi_{l,j,i}, \quad l = 0, \cdots, L - 1, \quad i = 1, \cdots, n_l, \quad j = 1, \cdots, n_{l+1}
\end{aligned}
\label{eq:acti_funct}
\end{equation}
with $n_l$ is the number of nodes of the layer $l^{th}$. The calculation of input \(\mathbf{x}_l\) through the function matrix \(\Phi_{l}\) with size \(n_{l+1} \times n_l\) to compute the output \(\mathbf{x}_{l+1}\) at the \((l+1)^{\text{th}}\) layer is given as:

\begin{equation}
\begin{aligned}
\mathbf{x}_{l+1} = 
\underbrace{\left(
\begin{array}{cccc}
\phi_{l,1,1}(\cdot) & \phi_{l,1,2}(\cdot) & \cdots & \phi_{l,1,n_l}(\cdot) \\
\phi_{l,2,1}(\cdot) & \phi_{l,2,2}(\cdot) & \cdots & \phi_{l,2,n_l}(\cdot) \\
\vdots & \vdots & \ddots & \vdots \\
\phi_{l,n_{l+1},1}(\cdot) & \phi_{l,n_{l+1},2}(\cdot) & \cdots & \phi_{l,n_{l+1},n_l}(\cdot)
\end{array}\right)}_{\Phi_{l}} \mathbf{x}_l
\label{eq:function_matrix}
\end{aligned}
\end{equation}

\subsection{Implementation of the Current KANs}


\begin{figure*}[htbp]
  \centering
\includegraphics[scale=0.95]{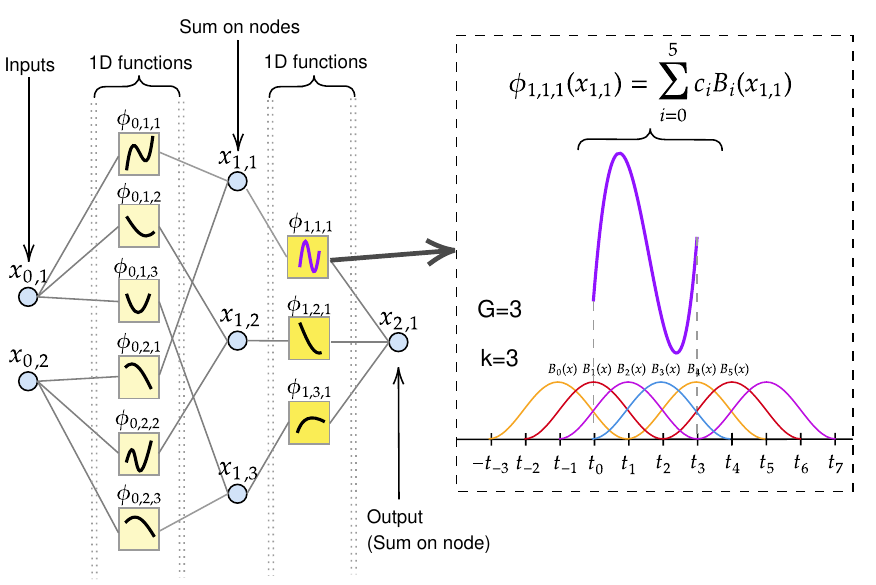}
  \centering
  \caption{Left: The structure of KAN(2,3,1). Right: The simulation of how to calculate $\phi_{1,1,1}$ by control points and B-splines~\cite{ta2024fc}. \(G\) and \(k\) is the grid size and the spline order, the number of B-splines equals \(G + k = 3 + 3 = 6\).}
\label{fig:kan_diagram}
\end{figure*}


LiuKAN\footnote{We refer to the original KAN as LiuKAN, named after the first author's last name~\cite{liu2024kan}, while another study~\cite{bozorgasl2024wav} calls it Spl-KAN.}, as implemented by \citet{liu2024kan}, is considered the original KAN design. It uses a residual activation function \(\phi(x)\), which is defined as the sum of a base function and a spline function, with their respective weight matrices \(w_b\) and \(w_s\).
\begin{equation}
\begin{aligned}
\phi(x) = w_b b(x) + w_s spline(x)
\end{aligned}
\label{eq:acti_funct_imp}
\end{equation}

\begin{equation}
\begin{aligned}
b(x) = silu(x) = \frac{x}{1 + e^{-x}}
\end{aligned}
\label{eq:b_function}
\end{equation}

\begin{equation}
\begin{aligned}
spline(x) = \sum_{i}c_iB_i(x)
\end{aligned}
\label{eq:spline_function}
\end{equation}
In \Cref{eq:acti_funct_imp}, \( b(x) \) is defined as \( silu(x) \) (as shown in \Cref{eq:b_function}), and \( spline(x) \) is represented as a linear combination of B-splines \( B_i \) and their respective control points (coefficients) \( c_i \) (as depicted in \Cref{eq:spline_function}). The activation functions are activated using \( w_s = 1 \), with \( spline(x) \approx 0 \), while \( w_b \) is initialized using Xavier initialization. Note that one may consider using another initialization method, such as Kaiming, or a custom version tailored to the specific problem~\cite{yang2024kolmogorov}. 

\Cref{fig:kan_diagram} illustrates the structure of KAN(2,3,1), which consists of 2 input nodes, 3 hidden nodes, and 1 output node. The result of each node is the summation of single functions \(\phi\), which are described as "edges". It also demonstrates how to compute the inner function \(\phi\) using control points (coefficients) and B-splines. The number of B-splines is calculated as the sum of the grid size \(G\) and the spline order \(k\), so we have \( G + k = 3 + 3 = 6 \), corresponding to the range of $i$ is from 0 to 5.

\textbf{EfficientKAN} follows a similar approach to \citet{liu2024kan}, but it reworks the computation using B-splines and linear combinations, which helps reduce memory usage and simplifies the calculations~\cite{Blealtan2024}. The authors replaced the previous L1 regularization on input samples with L1 regularization applied to the weights. They also introduced learnable scaling factors for the activation functions and switched the initialization of the base weight and spline scaler matrices to Kaiming uniform initialization, leading to significant improvements in performance on the MNIST dataset.

\textbf{FastKAN} can speed up training compared to EfficientKAN by utilizing RBFs to approximate the 3rd-order B-spline and incorporating layer normalization to maintain inputs within the RBFs' domain~\cite{li2024kolmogorov}. These modifications simplify the implementation without compromising accuracy. The RBF has the following formula:

\begin{equation}
\begin{aligned}
\phi(r) = e^{-\epsilon r^2}
\end{aligned}
\label{eq:gaussian_rbf}
\end{equation}

The distance between an input vector \( x \) and a center \( c \) is denoted by \( r = \|x - c\| \), and \( \epsilon \) (where \( \epsilon > 0 \)) is a parameter that adjusts the width of the Gaussian function. In FastKAN, Gaussian Radial Basis Functions (GRBFs) are used, with \( \epsilon = \frac{1}{2h^2} \), as described in~\cite{li2024kolmogorov}, given by:

\begin{equation}
\begin{aligned}
\phi_{\mathit{RBF}}(r) = \exp\left(-\frac{r^2}{2h^2}\right)
\end{aligned}
\label{eq:special_gaussian_rbf}
\end{equation}
and \( h \) controls the width of the Gaussian function. Then, the RBF network with \( C \) centers can be represented as~\cite{li2024kolmogorov,ta2024bsrbf,ta2024fc}:

\begin{equation}
\begin{aligned}
RBF(x) = \sum_{i=1}^{C} w_i \phi_{\mathit{RBF}}(r_i) = \sum_{i=1}^{C} w_i \exp\left(-\frac{||x - c_i||}{2h^2}\right)
\end{aligned}
\label{eq:rbf_network}
\end{equation}
The weight \( w_i \) corresponds to the adjustable coefficients, and \( \phi \) symbolizes the radial basis function, as described in \Cref{eq:gaussian_rbf}.

FasterKAN demonstrates superior performance compared to FastKAN in both forward and backward processing speeds~\cite{athanasios2024}. It employs Reflectional Switch Activation Functions (RSWAFs), which are simplified versions of RBFs and easily computable thanks to their uniform grid configuration. The RSWAF function is expressed as:

\begin{equation}
\begin{aligned}
\phi_{\mathit{RSWAF}}(r) = 1 - \left(\tanh\left(\frac{r}{h}\right)\right)^2
\end{aligned}
\label{eq:rswaf_funct}
\end{equation}

The RSWAF network with $N$ centers can be represented as:
\begin{equation}
\begin{aligned}
\mathit{RSWAF}(x) = \sum_{i=1}^{N} w_i \phi_{\mathit{RSWAF}}(r_i) = \sum_{i=1}^{N} w_i \left(1 - \left(\tanh\left(\frac{||x - c_i||}{h}\right)\right)^2\right)
\end{aligned}
\label{eq:rswaf_network}
\end{equation}

\textbf{BSRBF-KAN} is a KAN that integrates B-splines from EfficientKAN and GRBFs from FastKAN in each layer, using additions~\cite{ta2024bsrbf}. It performs faster convergence compared to EfficientKAN, FastKAN, and FasterKAN during training. However, this property may lead to overfitting and does not guarantee a high validation accuracy. The BSRBF function is expressed as:

\begin{equation}
\begin{aligned}
\phi_{BSRBF}(x)  =  w_b b(x) + w_s (\phi_{BS}(x) + \phi_{RBF}(x)) 
\end{aligned}
\label{eq:bsrbf_funct}
\end{equation}
The base function \( b(x) \) along with its corresponding matrix \( w_b \) represents the linear component of traditional MLP layers, while \( \phi_{BS}(x) \) and \( \phi_{RBF}(x) \) denote the B-spline and Radial Basis Function (RBF) respectively. The matrix \( w_s \) refers to the spline-related coefficients.

\subsection{Parameter Requirements in KANs vs. MLPs}

Consider an input \( x \), along with a network layer characterized by an input dimension \( d_{\text{in}} \) and an output dimension \( d_{\text{out}} \). Let \( k \) represent the spline order and \( G \) denote the grid size of a function, such as B-splines, utilized in Kolmogorov-Arnold Networks (KANs).
The number of control points (also the number of basis functions) required is \( G + k \). The total number of parameters, including the weight matrix and the bias term, when passing \( x \) through a KAN layer, is:

\begin{equation}
\begin{aligned}
KAN_{\text{params}} = 
\underbrace{d_{in} \times d_{out} \times (G + k)}_{\text{weight matrix params}} 
+ 
\underbrace{d_{out}}_{\text{bias matrix params}}
\end{aligned}
\label{eq:kan_params}
\end{equation}
while an MLP layer only needs:

\begin{equation}
\begin{aligned}
MLP_{\text{params}} = 
\underbrace{d_{in} \times d_{out}}_{\text{weight matrix params}} 
+ 
\underbrace{d_{out}}_{\text{bias matrix params}}
\end{aligned}
\label{eq:mlp_params}
\end{equation}
From \Cref{eq:kan_params} and \Cref{eq:mlp_params}, it is clear that KANs always use more parameters than MLPs, making comparisons with MLPs in networks with the same layer structure unfair.

Say that 4 given points are combined as 1 batch; therefore, we have input data \((1, 4)\). Also, say \(d_{in}\) and \(d_{out}\) of a network layer are 4 and 1. According to \Cref{eq:kan_params} and \Cref{eq:mlp_params}, the number of parameters produced by an MLP layer with the Sigmoid function (or any activation function) equals \(4 \times 1 + 4 = 8\). For B-splines, suppose that the degree of the function is \(k = 3\) and the grid size is \(G = 5\). Note that these values are standard in many existing KANs. In this case, the number of parameters for a KAN layer is calculated as \(4 \times 1 \times (3 + 5) + 4 = 36\). Therefore, it is not surprising that a KAN captures more data features than an MLP, as it uses more parameters.

In another example, given a list with 4 data points, \([0.4, 0.5, 0.6, 0.7]\), when passing this list to the Sigmoid function ($\sigma(x) = 1/(1 + e^{-x})$), the output is \(\textit{tensor}\left( \begin{bmatrix} \begin{bmatrix} 0.5986, 0.6224, 0.6456, 0.6681 \end{bmatrix} \end{bmatrix} \right)\), with shape \((1, 4)\). When doing this with a B-spline in EfficientKAN\footnote{The output has the same features (shape and partition of unity) when using Liu-KAN.}, the output is as follows:

\[
\textit{tensor}\left(
\begin{bmatrix}
\begin{bmatrix}
\begin{bmatrix}0.0000, 0.0000, 0.0000, 0.0208, 0.4792, 0.4792, 0.0208, 0.0000\end{bmatrix}, \\
\begin{bmatrix}0.0000, 0.0000, 0.0000, 0.0026, 0.3151, 0.6120, 0.0703, 0.0000 \end{bmatrix},\\
\begin{bmatrix}0.0000, 0.0000, 0.0000, 0.0000, 0.1667, 0.6667, 0.1667, 0.0000 \end{bmatrix},\\
\begin{bmatrix}0.0000, 0.0000, 0.0000, 0.0000, 0.0703, 0.6120, 0.3151, 0.0026 \end{bmatrix} \

\end{bmatrix}
\end{bmatrix}
\right)
\]
This tensor has a shape of $(1, 4, 8)$, where the batch size is 1, 4 data points are processed, and 8 basis functions  ($G + k$ = 5 + 3) are evaluated per input. Each row sums approximately to 1 due to the partition of unity property of B-spline basis functions, which ensures smoothness, locality, and a complete representation of the input. 

\begin{figure*}[htbp]
  \centering
\includegraphics[scale=0.55]{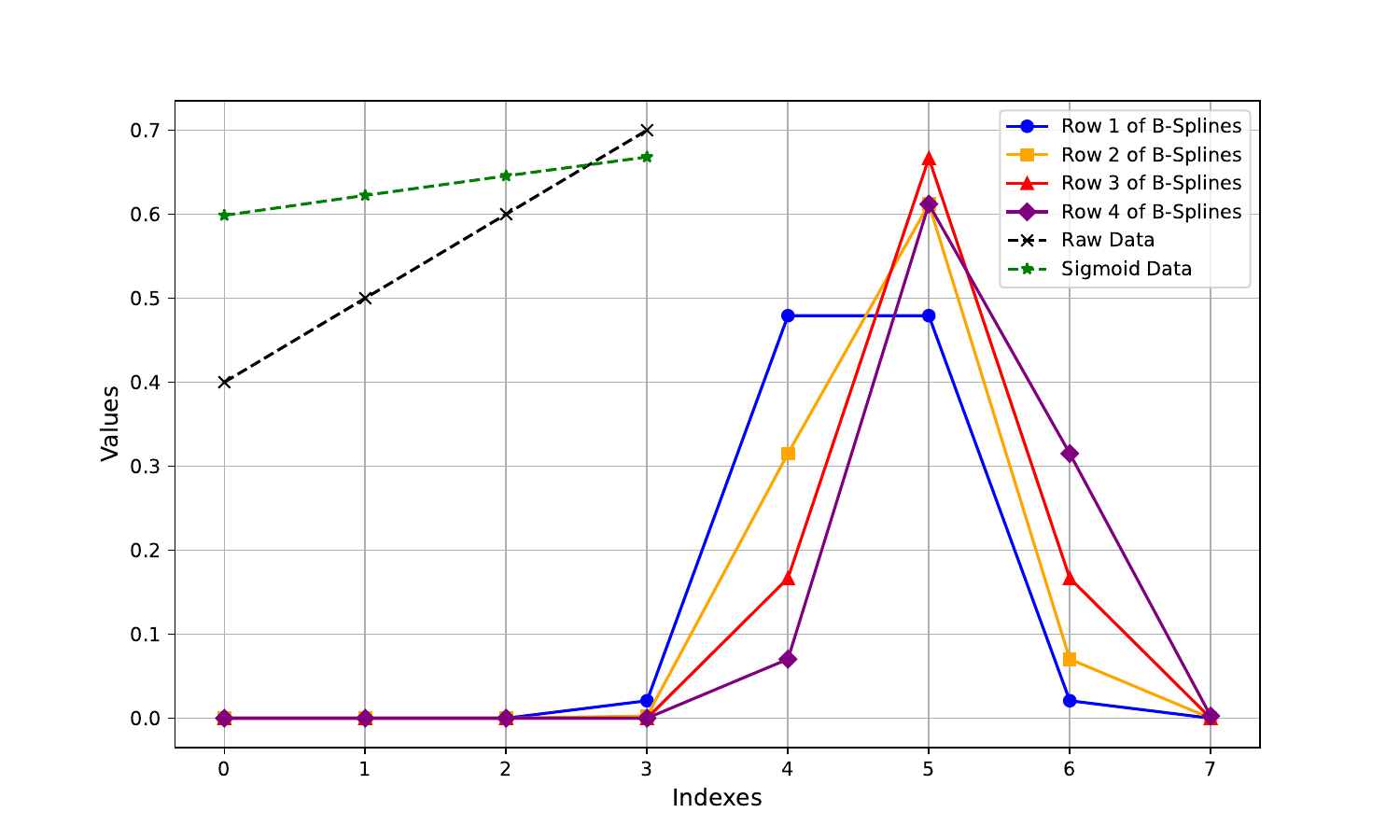}
  \centering
  \caption{Plots of outputs generated by Sigmoid and B-spline functions. The more B-splines are used, the smoother the curves become.}
\label{fig:example_plot}
\end{figure*}

B-splines are localized, meaning that only a few basis functions contribute significantly to the result at any given point~\cite{barsky1983local,liu2019splinets}. They often display symmetry, resulting in a bell-shaped distribution that depends on the placement of knots, as shown in \Cref{fig:example_plot}. Traditional activation functions like the sigmoid produce a single output for each input, whereas B-spline functions generate multiple outputs through their basis function representation. This enables B-splines to capture data features more comprehensively.

\subsection{PRKAN}

Given a set of inputs with (batch) size \( B \), each with a data dimension \( D \) (\( D = W \times H \), where \( W \) is the width and \( H \) is the height), the input tensor \( X \) has a shape of \( (B, D) \). We experiment with black-and-white image data, where the number of channels is \( T = 1 \). Thus, we do not explicitly consider \( T \) in this work; however, the same principles can be extended to multi-channel data. Let \( d_{\text{in}} \) and \( d_{\text{out}} \) represent the input and output dimensions of a layer, respectively. Note that \( d_{\text{in}} \) is always equal to \( D \) in each layer to ensure consistency throughout the network. So, when \( X \) passes through this layer, the output \( Y \) has a shape of \( (B, d_{\text{out}}) \).

When passing through an MLP layer, the layer output \( Y_{\text{MLP}} \) has shape \( (B, d_{\text{out}}) \) after performing a linear transformation by multiplying the weight matrix of shape \( (d_{\text{in}}, d_{\text{out}}) \) with the input tensor of shape \( (B, D) \), then adding the bias matrix, followed by the application of an activation function. Therefore, an MLP layer first performs the linear transformation, followed by the activation function. Sometimes, this order can be reversed, with the activation function applied first, followed by the linear transformation. In this case, we first apply the activation function to the output tensor of shape \( (B, D) \), then multiply it by the weight matrix of shape \( (d_{\text{in}}, d_{\text{out}}) \) and add the bias matrix to obtain the layer output. We call the method of data transformation in an MLP layer \texttt{base}.

For a KAN layer, the spline tensor \( X_{\text{spline}} \) has a shape of \( (B, D, G + k) \) when passing \(X\) to the B-spline functions. Note that when using RBFs, we can also specify the number of centers, denoted as $C$. For a fair comparison between B-splines and RBFs, $C$ is equal to \( C = G + k \). Then, the linear transformation is then performed by multiplying it with the weight matrix of the shape \( (D \times (G + k), d_{\text{out}}) \) and adding the bias matrix to obtain the layer output \( Y_{\text{KAN}} \), which has the same shape as \( Y_{\text{MLP}} \), i.e., \( (B, d_{\text{out}}) \). The problem with KAN layers is that their weight matrices contain significantly more parameters than those of MLP layers. To ensure the approximate number of parameters remains the same as that of the MLP layer, we can \textbf{convert \( X_{\text{spline}} \) with a shape of \( (B, D, G + k) \) to \( (B, D) \)}, so that we can then apply a linear transformation by multiplying it with a weight matrix of shape \( (d_{\text{in}}, d_{\text{out}}) \) and adding a bias matrix to obtain the layer output \( (B, d_{\text{out}}) \). 

\begin{figure*}[htbp]
  \centering
\includegraphics[scale=0.8]{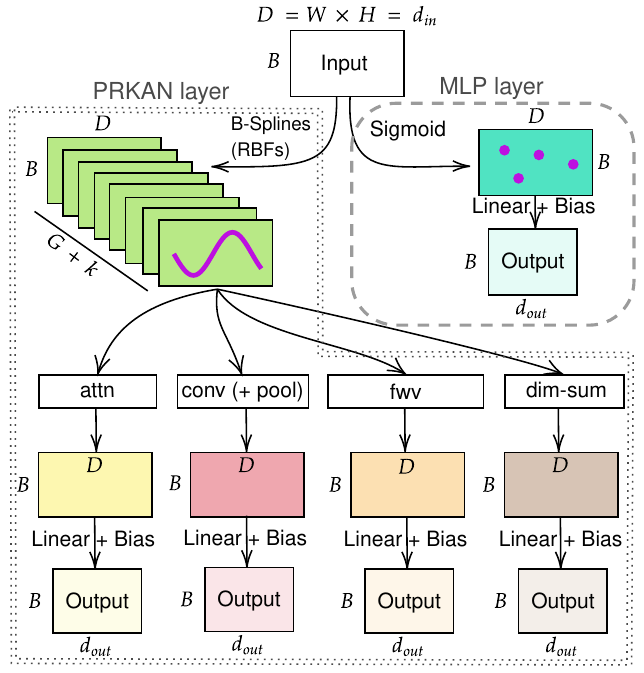}
  \centering
  \caption{The diagram illustrates how the input \( (B, D) \) is passed through both MLP and PRKAN layers to produce the output \( (B, d_{\text{out}}) \). Convolutional layers can be used independently or in combination with pooling layers to reduce data dimensionality.}
\label{fig:mlp_prkan_layer}
\end{figure*}

\begin{figure*}[htbp]
  \centering
\includegraphics[scale=0.9]{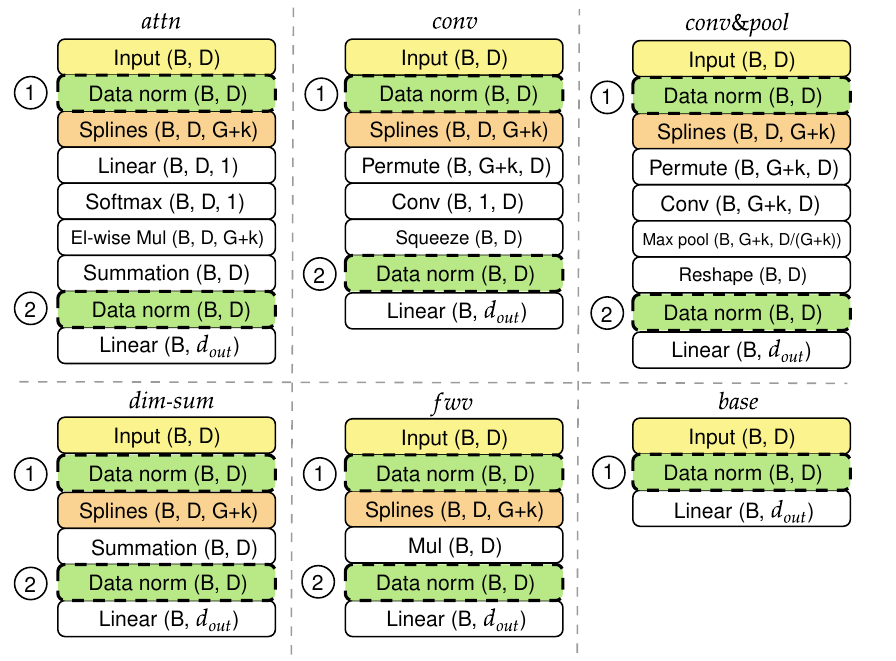}
  \centering
  \caption{The architectures are defined by methods in PRKANs (\texttt{attn}, \texttt{conv}, \texttt{conv\&pool}, \texttt{dim-sum}, and \texttt{fwv}) and MLPs (\texttt{base}), along with suggestions on where to apply data normalization. Two key positions (1 \& 2) are proposed for applying data normalization on tensor data with the shape $(B, D)$.}
\label{fig:prkan_data_norm}
\end{figure*}

Various methods exist to reduce the dimensionality of a data tensor. However, selecting methods that optimize computational efficiency and maintain competitive performance during the training process of neural networks is of paramount importance. Therefore, we propose several methods to reduce the use of parameters in KAN layers as MLP layers: attention mechanisms (\texttt{attn}), convolutional layers (\texttt{conv}), convolutional layers + pooling layers (\texttt{conv\&pool}), dimension summation (\texttt{dim-sum}), and feature weight vectors (\texttt{fwv}), as shown in \Cref{fig:mlp_prkan_layer}. For a glance at the detailed architecture without delving into the equations of methods, please refer directly to \Cref{fig:prkan_data_norm}. 

By combining all these methods, we introduce a novel KAN model named \textbf{PRKAN} (\textbf{P}arameter-\textbf{R}educed \textbf{K}olmogorov-\textbf{A}rnold \textbf{N}etwork). PRKANs strictly adhere to the same number of parameters and network structure, ensuring fair comparability with MLPs. To the best of our knowledge, PRKANs may also be one of the first KANs to achieve this. Note that we do not maintain the same FLOPs (Floating Point Operations) for PRKANs as MLPs, as this would make the design of the PRKAN architecture more challenging and complex.

\subsubsection{Attention Mechanism}


Attention mechanisms enhance neural networks by dynamically emphasizing relevant input features, leading to improved accuracy and interpretability. Inspired by global attention~\cite{liu2021global}, we apply multiplicative attention with a softmax-based weighting scheme along the data dimension \( D \) to ensure all input positions influence the final weighted sum.

For a spline data \( X_{\text{spline}} \) with shape \( (B, D, G + k) \), a linear transformation is applied to reduce the size \( G + k \) to 1. A softmax function is then applied along the \( D \) dimension to calculate the attention weights \( W_{\text{att}} \) with shape \( (B, D, 1) \). Next, \( X_{\text{spline}} \) is multiplied element-wise by \( W_{\text{att}} \), and the resulting tensor is summed over the last dimension to produce data with shape \( (B, D) \). Finally, a linear transformation is applied by multiplying with a weight matrix and adding a bias to produce the output \(Y_{\text{out}}\) of shape \( (B, d_{\text{out}}) \).

\begin{subequations}
\begin{equation}
    \text{Spline tensor: } \\
    X_{\text{spline}} \in \mathbb{R}^{B \times D \times (G + k)} \label{eq:input_rep_attn}
\end{equation}

\begin{equation}
    \text{Linear transformation: } \\
    X_{\text{linear}} = W_{\text{linear}} \times X_{\text{spline}} + b_{\text{linear}}, \quad X_{\text{linear}} \in \mathbb{R}^{B \times D \times 1} \label{eq:linear_transformation_attn}
\end{equation}

\begin{equation}
    \text{Softmax over the data dimension $D$: } \\
    W_{\text{att}} = \textit{softmax}(X_{\text{linear}}, \text{dim}=-2), \quad W_{\text{att}} \in \mathbb{R}^{B \times D \times 1} \label{eq:softmax_attn}
\end{equation}

\begin{equation}
    \text{Element-wise multiplication: } \\
    X' = X_{\text{spline}} \odot W_{\text{att}}, \quad X' \in \mathbb{R}^{B \times D \times (G + k)} \label{eq:elementwise_multiplication_attn}
\end{equation}

\begin{equation}
    \text{Summation along the last dimension: } \\
    X'' = \sum_{\text{dim}=-1} X', \quad X'' \in \mathbb{R}^{B \times D} \label{eq:summation_attn}
\end{equation}

\begin{equation}
    \text{Linear transformation: } \\
    X_{\text{out}} = W_{\text{out}} \times \sigma (X'')   + b_{\text{out}}, \quad X_{\text{out}} \in \mathbb{R}^{B \times d_{\text{out}}} \label{eq:final_linear_attn}
\end{equation}
\end{subequations}

\subsubsection{Convolution Layers}
Convolutional layers are used to extract features from data in the spatial domain, and they can also be utilized to reduce data dimensions. For a given spline tensor \(X_{\text{spline}}\) of shape \((B, D, G + k)\), we first permute it into \((B, G + k, D)\), where \(G + k\) represents the number of input channels, and \(1\) is the number of output channels in a 1D convolution. A single convolutional layer is applied to reduce the input from \((B, G + k, D)\) to \((B, 1, D)\). The result is then squeezed to \((B, D)\). Finally, a linear transformation is applied by multiplying the data with a weight matrix and adding a bias term to produce an output \(X_{\text{out}}\) of shape \((B, d_{\text{out}})\).

\begin{subequations}
\begin{equation}
    \text{Spline tensor:} \quad X_{\text{spline}} \in \mathbb{R}^{B \times D \times (G+k)} \label{eq:input_tensor_conv}
\end{equation}

\begin{equation}
    \text{Permute tensor:} \quad X_{\text{perm}} = \textit{permute}(X_{\text{spline}}, 0, 2, 1), \quad X_{\text{perm}} \in \mathbb{R}^{B \times (G+k) \times D} \label{eq:permute_conv}
\end{equation}

\begin{equation}
    \text{1D convolution:} \quad X_{\text{conv}} = W_{\text{conv}}  \times X_{\text{perm}} + b_{\text{conv}}, \quad X_{\text{conv}} \in \mathbb{R}^{B \times 1 \times D} \label{eq:conv_conv}
\end{equation}

\begin{equation}
    \text{Squeeze dimension:} \quad X_{\text{squeeze}} =  \textit{squeeze}(X_{\text{conv}},1), \quad X_{\text{squeeze}} \in \mathbb{R}^{B \times D} \label{eq:squeeze_conv}
\end{equation}

\begin{equation}
    \text{Linear transformation:} \quad X_{\text{out}} = W_{\text{out}} \times \sigma (X_{\text{squeeze}}) + b_{\text{out}}, \quad X_{\text{out}} \in \mathbb{R}^{B \times d_{out}} \label{eq:linear_transformation_conv}
\end{equation}
\end{subequations}

\subsubsection{Convolution Layers + Pooling Layers}

For a given spline tensor \(X_{\text{spline}}\) of shape \((B, D, G + k)\), we first permute it into \((B, G + k, D)\), where \(G + k\) represents the number of input channels and is also the number of output channels in a 1D convolution. A single convolutional layer is applied that does not change the input shape, transforming it from \((B, G + k, D)\) to \((B, G + k, D)\). The 1x1 convolution combines and transforms features across same channels, enabling richer representations while preserving the spatial layout. Next, we use a max pooling layer with a kernel size of \(G + k\), reducing the tensor shape from \((B, G + k, D)\) to \((B, G + k, D / (G + k))\). After pooling, the tensor is reshaped from \((B, G + k, D / (G + k))\) to \((B, D)\). Finally, a linear transformation is applied by multiplying the data with a weight matrix and adding a bias term, producing an output \(X_{\text{out}}\) of shape \((B, d_{\text{out}})\).

\begin{equation}
    \text{Spline tensor:} \quad X_{\text{spline}} \in \mathbb{R}^{B \times D \times (G+k)} 
     \label{eq:input_tensor_conv_pool}
\end{equation}

\begin{equation}
    \text{Permute tensor:} \quad X_{\text{perm}} = \textit{permute}(X_{\text{spline}}, 0, 2, 1), \quad X_{\text{perm}} \in \mathbb{R}^{B \times (G+k) \times D} 
    \label{eq:permute_conv_pool}
\end{equation}

\begin{equation}
    \text{1D convolution:} \quad X_{\text{conv}} = W_{\text{conv}} \times X_{\text{perm}} + b_{\text{conv}}, \quad X_{\text{conv}} \in \mathbb{R}^{B \times (G+k) \times D} 
    \label{eq:conv_conv_pool}
\end{equation}

\begin{equation}
    \text{Max pooling:} \quad X_{\text{pool}} = \textit{pool}(X_{\text{conv}}), \quad X_{\text{pool}} \in \mathbb{R}^{B \times (G+k) \times \frac{D}{G+k}}
     \label{eq:pool_conv_pool}
\end{equation}

\begin{equation}
    \text{Reshape:} \quad X_{\text{reshaped}} = \textit{reshape}(X_{\text{pool}}), \quad X_{\text{reshaped}} \in \mathbb{R}^{B \times D}
    \label{eq:reshape_conv_pool}
\end{equation}

\begin{equation}
    \text{Linear transformation:} \quad X_{\text{out}} = W_{\text{out}} \times X_{\text{reshaped}} + b_{\text{out}}, \quad X_{\text{out}} \in \mathbb{R}^{B \times d_{\text{out}}}
    \label{eq:linear_conv_pool}
\end{equation}

\subsubsection{Dimension Summation}
This method is the simplest and fastest in practice. It only requires operations such as summation along the last data dimension. For a spline tensor \(X_{\text{spline}}\) of shape \((B, D, G + k)\), we first apply summation over the \(G + k\) dimension to obtain a tensor of shape \((B, D)\). Next, a linear transformation is performed on the resulting tensor to produce an intermediate output of the same shape, \((B, D)\). Finally, another linear transformation is applied by multiplying the intermediate output with a weight matrix and adding a bias term, resulting in an output tensor \(X_{\text{out}}\) of shape \((B, d_{\text{out}})\).

\begin{subequations}
\begin{equation}
    \text{Spline tensor:} \quad X_{\text{spline}} \in \mathbb{R}^{B \times D \times (G+k)} \label{eq:spline_input_dimsum}
\end{equation}

\begin{equation}
    \text{Summation along the last dimension: } \\
    X' = \sum_{\text{dim}=-1} X_{\text{spline}}, \quad X' \in \mathbb{R}^{B \times D} \label{eq:summation_dimsum}
\end{equation}

\begin{equation}
    \text{Linear transformation:} \quad X_{\text{out}} = W_{\text{out}} \times \sigma(X') + b_{\text{out}}, \quad X_{\text{out}} \in \mathbb{R}^{B \times d_{\text{out}}} \label{eq:final_linear_dimsum}
\end{equation}
\end{subequations}

\subsubsection{Feature Weight Vectors}

Since we have the spline tensor \( X_{\text{spline}} \) with shape \( (B, D, G + k) \) after a KAN layer, we might consider multiplying it with a learnable feature vector of shape \( (G + k, 1) \), based on the feature dimension \( (G + k) \), to reduce the data tensor shape to \( (B, D) \). Then, we can apply a linear transformation by multiplying the intermediate output with a weight matrix and adding a bias term, resulting in an output tensor \( X_{\text{out}} \) of shape \( (B, d_{\text{out}}) \).

\begin{subequations}
\begin{equation}
    \text{Spline tensor:} \quad X_{\text{spline}} \in \mathbb{R}^{B \times D \times (G+k)} \label{eq:input_tensor_fwv}
\end{equation}

\begin{equation}
    \text{Multiplying with feature weight vector:} \quad X' = X_{\text{spline}} \times M, \quad M \in \mathbb{R}^{(G+k) \times 1}, \quad X' \in \mathbb{R}^{B \times D} \label{eq:multiplication_fwv}
\end{equation}

\begin{equation}
    \text{Linear transformation:} \quad X_{\text{out}} = W_{\text{out}} \times \sigma(X') + b_{\text{out}}, \quad X_{\text{out}} \in \mathbb{R}^{B \times d_{\text{out}}} \label{eq:final_linear_fwv}
\end{equation}
\end{subequations}

\subsubsection{Data Normalization}
\label{data_norm}
Although data normalization offers various advantages in training neural networks, our main focus is on its ability to speed up convergence and enhance accuracy. Therefore, we implement two commonly used forms of data normalization: batch normalization~\cite{ioffe2015batch} and layer normalization~\cite{ba2016layer}. Depending on the specific properties of dimension reduction methods, the most effective positions for applying data normalization to achieve optimal results will vary.

\Cref{fig:prkan_data_norm} illustrates the recommended positions (1 \& 2) for data normalization in PRKAN methods. To simplify the design of PRKANs, data normalization should be applied either immediately after the input or following any intermediate representations that retain the same data shape \((B, D)\) as the input. To reduce computational overhead, we apply data normalization at a single position, as multiple positions are not required. While normalization can be applied to the output, we prefer not to do so because the output of one layer serves as the input to the next layer, which also applies normalization to its input. The model performance details based on data normalization positions is presented in \Cref{norm_positions}, \Cref{tab:norm_pos_mnist}, and  \Cref{tab:norm_pos_fashion_mnist}.

\section{Experiments}
\subsection{Training Configuration}
\label{sec:training_conf}

We evaluated PRKANs using parameter reduction methods and compared their performance with MLPs in the MNIST~\cite{deng2012mnist} and Fashion-MNIST~\cite{xiao2017fashion} datasets. Each model is trained over 5 independent runs, with average values for metrics such as training accuracy, validation accuracy, F1 score, and training time to minimize variability. All models share the same architecture with 784 input neurons, 64 hidden neurons, and 10 output neurons corresponding to the 10 output classes (0–9), arranged as (784, 64, 10). We incorporate B-splines from EfficientKAN and GRBFs from FastKAN, with slight modifications to enhance data feature capture. For PRKANs, we select GRBFs and SiLUs as the default functions—GRBFs for their speed (faster than B-splines~\cite{li2024kolmogorov}) and SiLUs for their ability to promote smooth and effective gradient propagation. Furthermore, the code implementations of previous works~\cite{liu2024kan,liu2024kan2.0,ta2024bsrbf,ta2024fc,bozorgasl2024wav} on KANs primarily focused on SiLU, motivating us to make a fair comparison with them in the experiments. 

\begin{table*}[ht]
	\caption{The number of used parameters and FLOPs by models.}
	\centering
	\begin{tabular}{p{2.5cm}p{3cm}p{3cm}p{2.5cm}p{2cm}}
            \hline
		\textbf{Dataset} & \textbf{Model} & \textbf{Network structure} & \textbf{\#Used Params} & \textbf{\#FLOPs} \\
    \hline 
    \multirow{10}{2.5cm}{\textbf{MNIST \\ \& \\ Fashion-MNIST}}  
& PRKAN-attn	& (784, 64, 10) & 52604 & 20.36K  \\
& PRKAN-conv	& (784, 64, 10) & 52604 & 18.66K \\
& PRKAN-conv\&pool	& (784, 64, 10) & 52730 & \textbf{252.72K} \\
& PRKAN-dim-sum & (784, 64, 10) & 52586 & 3.4K \\
& PRKAN-fwv & (784, 64, 10) & 52602 & 16.96K \\
& MLP-base & (784, 64, 10) & 52512 & 1.844K \\
& BSRBF-KAN & (784, 7, 10) & 51588 & 3.16K \\
& EfficientKAN & (784, 7, 10) & \textbf{55580} & \textbf{1.582K} \\
& FastKAN & (784, 7, 10) & 51605 & 12.74K \\
& FasterKAN & (784, 8, 10) & 52382 & 26.92K \\
            \hline
            \multicolumn{5}{l}{\texttt{attn} = attention mechanism, \texttt{conv} = convolutional layers, \texttt{conv\&pool} = convolutional \& pooling layers}  \\
              \multicolumn{5}{l}{\texttt{dim-sum} = dimension summation, \texttt{fwv} = feature weight vectors, \texttt{base} = MLP layers} \\
             \hline
	\end{tabular}
	\label{tab:model_params_flops}
\end{table*}

The training process spans 25 epochs for MNIST and 35 epochs for Fashion-MNIST, balancing convergence with computational efficiency. We apply a consistent set of hyperparameters across all experiments, including \texttt{grid\_size=5} (as $G$ in B-splines), \texttt{spline\_order=3} (as $k$ in B-splines), and \texttt{num\_grids=8} (as $C$ in RBFs), to ensure comparability. Other settings include \texttt{batch\_size=64}, \texttt{learning\_rate=1e-3}, \texttt{weight\_decay=1e-4}, \texttt{gamma=0.8}, \texttt{optimizer=AdamW}, and \texttt{loss=CrossEntropy}. These hyperparameters were selected based on preliminary tuning experiments to optimize model performance across datasets.

\Cref{tab:model_params_flops} shows the number of parameters and FLOPs used by methods in PRKANs and MLPs. We use the \texttt{ptflops}\footnote{\url{https://pypi.org/project/ptflops/}} package to calculate the number of MACs (multiply–accumulate operations), then multiply by 2 to obtain the number of FLOPs. While parameter-reduction methods help keep the number of parameters comparable to MLPs, their FLOPs are generally higher, as they often introduce more complex operations to reduce the model's size. Especially \texttt{PRKAN-conv\&pool}, whose FLOPs are more than 137 times higher than MLPs due to the additional convolution and pooling operations. On the other hand, \texttt{PRKAN-dim-sum} has the least number of FLOPs among the PRKAN methods, with a value more than 1.84 times higher than that of MLPs. Besides, we modify the network structure of some existing KANs to maintain an approximately equal number of parameters compared to MLPs and PRKANs. While EfficientKAN has slightly more parameters than others, it boasts the fewest FLOPs, followed by BSRBF-KAN and \texttt{PRKAN-dim-sum}.

\subsection{PRKANs vs. MLPs}

\begin{table*}[ht]
	\caption{The comparison between PRKANs and MLPs. We select the best model performance based on data normalization positions (1 or 2) as structured in \Cref{fig:prkan_data_norm} and experimented in \Cref{norm_positions}, \Cref{tab:norm_pos_mnist}, and \Cref{tab:norm_pos_fashion_mnist}.}
	\centering
	\begin{tabular}{p{1.5cm}p{3cm}p{1.5cm}p{1.8cm}p{1.8cm}p{1.8cm}p{1.8cm}}
            \hline
		\textbf{Dataset} &  \textbf{Model}  & \textbf{Norm.} & \textbf{Train. Acc.} & \textbf{Val. Acc.} & \textbf{F1} & \textbf{Time (sec)} \\
    \hline 
    \multirow{18}{1.5cm}{\textbf{MNIST}} 
    & PRKAN-attn & batch & 98.97 ± 0.33 & 97.29 ± 0.10 & 97.25 ± 0.10 & 179.35 \\
    & PRKAN-conv & batch & 98.51 ± 0.58 & 95.62 ± 0.30 & 95.56 ± 0.30 & 199.73 \\
    & PRKAN-conv\&pool & batch & 97.22 ± 0.42 & 94.52 ± 0.44 & 94.45 ± 0.44 & 196.03 \\
    & PRKAN-dim-sum & batch & 99.19 ± 0.15 &  95.54 ± 0.06 & 95.50 ± 0.06 & 180.51  \\
    & PRKAN-fwv & batch & 98.94 ± 0.62 & 94.74 ± 0.97 & 94.66 ± 0.99 & 171.98 \\
    & MLP-base & batch & 98.72 ± 0.23 & 95.95 ± 0.05 & 95.91 ± 0.05 & 165.84 \\
    & \cellcolor[gray]{0.95} PRKAN-attn & \cellcolor[gray]{0.95} layer & \cellcolor[gray]{0.95} 99.81 ± 0.09 & \cellcolor[gray]{0.95} \textbf{97.46 ± 0.06} & \cellcolor[gray]{0.95} \textbf{97.42 ± 0.06} & \cellcolor[gray]{0.95} 178.02 \\
    & \cellcolor[gray]{0.95} PRKAN-conv & \cellcolor[gray]{0.95} layer & \cellcolor[gray]{0.95} \textbf{99.85 ± 0.09} & \cellcolor[gray]{0.95} 97.26 ± 0.10 & \cellcolor[gray]{0.95} 97.23 ± 0.10 & \cellcolor[gray]{0.95} 193.07 \\
    & \cellcolor[gray]{0.95} PRKAN-conv\&pool & \cellcolor[gray]{0.95} layer & \cellcolor[gray]{0.95} 99.47 ± 0.10 & \cellcolor[gray]{0.95} 96.65 ± 0.23 & \cellcolor[gray]{0.95} 96.61 ± 0.24 & \cellcolor[gray]{0.95} 196.5 \\
    & \cellcolor[gray]{0.95} PRKAN-dim-sum & \cellcolor[gray]{0.95} layer & \cellcolor[gray]{0.95}  95.71 ± 0.12 & \cellcolor[gray]{0.95} 94.92 ± 0.13 & \cellcolor[gray]{0.95} 94.83 ± 0.14 & \cellcolor[gray]{0.95} 175.03 \\
    & \cellcolor[gray]{0.95} PRKAN-fwv & 
    \cellcolor[gray]{0.95} layer & \cellcolor[gray]{0.95} 99.58 ± 0.12 & \cellcolor[gray]{0.95} 97.19 ± 0.09 & \cellcolor[gray]{0.95} 97.15 ± 0.09 & \cellcolor[gray]{0.95} 168.92 \\
    & \cellcolor[gray]{0.95} MLP-base & \cellcolor[gray]{0.95} layer & \cellcolor[gray]{0.95} 99.84 ± 0.04 & \cellcolor[gray]{0.95} \textbf{97.72 ± 0.05} & \cellcolor[gray]{0.95} \textbf{97.69 ± 0.05} & \cellcolor[gray]{0.95} 162.58 \\
     & PRKAN-attn & none & 75.31 ± 5.52 & 75.36 ± 5.67 & 73.49 ± 6.36 & 176.45 \\
     & PRKAN-conv & none & 99.66 ± 0.07 & 97.12 ± 0.10 & 97.08 ± 0.10 & 179.13 \\
     & PRKAN-conv\&pool & none & 98.85 ± 0.11 & 95.79 ± 0.07 & 95.75 ± 0.06 & 183.78 \\
     & PRKAN-dim-sum & none & 10.92 ± 0.17 & 11.36 ± 0.00 & 2.04 ± 0.00 & 165.42 \\
     & PRKAN-fwv & none & 99.46 ± 0.14 & 97.01 ± 0.13 & 96.97 ± 0.13 & 165.07 \\
     & MLP-base  & none & 98.40 ± 0.04 & 97.38 ± 0.07 & 97.35 ± 0.07 & \textbf{156.68} \\
            \hline
            \hline
    \multirow{18}{1.5cm}{\textbf{Fashion-MNIST}} & PRKAN-attn & batch & 94.10 ± 0.17 & \textbf{88.87 ± 0.06} & \textbf{88.82 ± 0.06} & 259.18 \\ 
    & PRKAN-conv & batch & 94.23 ± 0.66 & 87.80 ± 0.25 & 87.73 ± 0.26 & 264.08 \\
    & PRKAN-conv\&pool & batch & 92.96 ± 1.18 & 87.61 ± 0.06 & 87.56 ± 0.06 & 278.91 \\
    & PRKAN-dim-sum & batch & 92.35 ± 0.09 & 87.06 ± 0.06 & 86.96 ± 0.06 & 244.03 \\
    & PRKAN-fwv & batch & 94.05 ± 0.75 & 87.79 ± 0.23 & 87.74 ± 0.23 & 245.86 \\
    & MLP-base & batch & 94.80 ± 0.06 & 88.29 ± 0.06 & 88.25 ± 0.06 & 233.71 \\
    & \cellcolor[gray]{0.95} PRKAN-attn & \cellcolor[gray]{0.95} layer & \cellcolor[gray]{0.95} 93.30 ± 0.20 & \cellcolor[gray]{0.95} 88.82 ± 0.09 & \cellcolor[gray]{0.95} 88.75 ± 0.10 & \cellcolor[gray]{0.95} 250.62 \\
    & \cellcolor[gray]{0.95} PRKAN-conv & \cellcolor[gray]{0.95} layer & \cellcolor[gray]{0.95} 93.76 ± 0.10 & \cellcolor[gray]{0.95}  88.47 ± 0.12 & \cellcolor[gray]{0.95} 88.43 ± 0.12 & \cellcolor[gray]{0.95}  260.52 \\
    & \cellcolor[gray]{0.95} PRKAN-conv\&pool & \cellcolor[gray]{0.95} layer & \cellcolor[gray]{0.95} \textbf{94.82 ± 0.42} & \cellcolor[gray]{0.95} 88.22 ± 0.07 & \cellcolor[gray]{0.95} 88.17 ± 0.08 & \cellcolor[gray]{0.95} 278.12 \\
    & \cellcolor[gray]{0.95} PRKAN-dim-sum & \cellcolor[gray]{0.95} layer & \cellcolor[gray]{0.95} 92.22 ± 0.05 & \cellcolor[gray]{0.95} 87.23 ± 0.04 & \cellcolor[gray]{0.95} 87.19 ± 0.05 & \cellcolor[gray]{0.95} 238.64 \\
    & \cellcolor[gray]{0.95} PRKAN-fwv & \cellcolor[gray]{0.95} layer & \cellcolor[gray]{0.95}  94.06 ± 0.27 & \cellcolor[gray]{0.95} 88.52 ± 0.08 & \cellcolor[gray]{0.95} 88.48 ± 0.08 & \cellcolor[gray]{0.95} 241.93 \\
    & \cellcolor[gray]{0.95} MLP-base & \cellcolor[gray]{0.95} layer & \cellcolor[gray]{0.95} 94.20 ± 0.09 & \cellcolor[gray]{0.95} \textbf{88.96 ± 0.05} & \cellcolor[gray]{0.95} \textbf{88.92 ± 0.05} & \cellcolor[gray]{0.95} 226.79 \\
    & PRKAN-attn & none & 59.35 ± 3.82 & 58.96 ± 3.78 & 54.15 ± 4.75 & 249.33 \\
    & PRKAN-conv & none & 92.65 ± 0.14 & 88.65 ± 0.06 & 88.58 ± 0.06 & 257.22 \\
    & PRKAN-conv\&pool & none & 94.08 ± 0.18 & 87.90 ± 0.08 & 87.87 ± 0.09 & 256.02 \\
    & PRKAN-dim-sum & none & 9.86 ± 0.06 & 10.02 ± 0.01 & 1.82 ± 0.00 & 236.59 \\
    & PRKAN-fwv & none & 92.82 ± 0.26 & 88.13 ± 0.23 & 88.07 ± 0.24 & 234.16 \\
    & MLP-base & none & 91.30 ± 0.05 & 88.33 ± 0.03 & 88.25 ± 0.02 & \textbf{221.25} \\
            \hline
             \multicolumn{7}{l}{Norm. = Data Normalization, Train. Acc = Training Accuracy, Val. Acc. = Validation Accuracy }  \\
             \multicolumn{7}{l}{\texttt{attn} = attention mechanism, \texttt{conv} = convolutional layers, \texttt{conv\&pool} = convolutional \& pooling layers}  \\
              \multicolumn{7}{l}{\texttt{dim-sum} = dimension summation, \texttt{fwv} = feature weight vectors, \texttt{base} = MLP layers}  \\
             
             \hline
	\end{tabular}
	\label{tab:prkans_vs_mlps}
\end{table*}

\begin{figure*}[htbp]
  \centering
\includegraphics[scale=0.55]{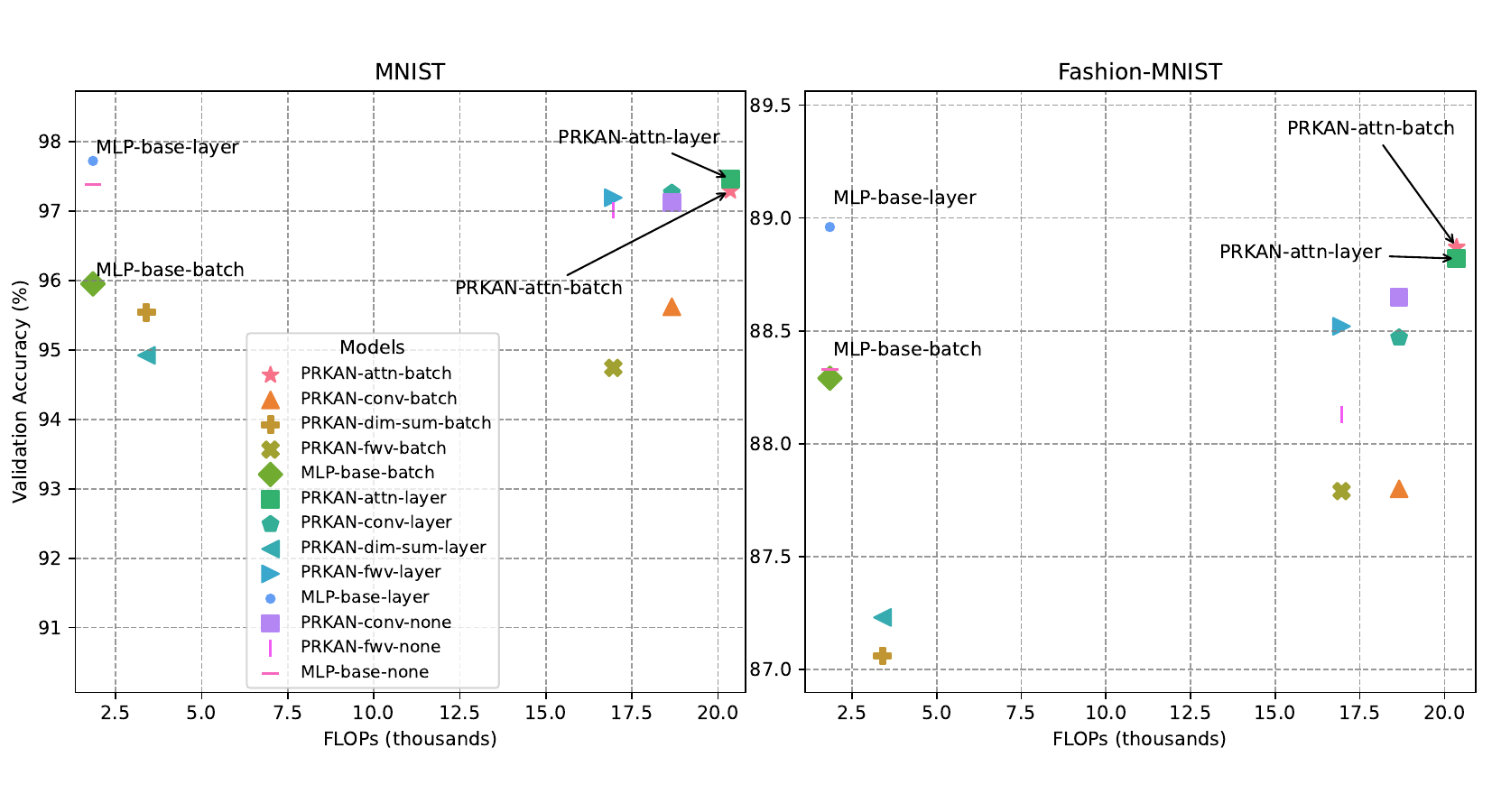}
  \centering
  \caption{Models by FLOPs and validation accuracy values. \texttt{PRKAN-conv\&pool} models are excluded due to their massive use of FLOPs. \texttt{PRKAN-attn} and \texttt{PRKAN-dim-sum} models without data normalization are also excluded due to their poor validation accuracy.}
\label{fig:flops_vs_val_acc}
\end{figure*}

The results in \Cref{tab:prkans_vs_mlps} indicate that PRKANs with attention mechanisms outperform other variants and can achieve performance comparable to that of MLPs. It also reveals that layer normalization consistently outperforms both batch normalization and the absence of any data normalization in PRKANs and MLPs. Among the evaluated models, \texttt{PRKAN-conv} and \texttt{PRKAN-conv\&pool} models with layer normalization demonstrate the best convergence during training, achieving average training accuracies of 99.85\% on MNIST and 94.82\% on Fashion-MNIST, respectively.

With batch normalization, \texttt{PRKAN-attn} models outperform MLPs and other PRKAN variants. Specifically, compared to MLPs, \texttt{PRKAN-attn} models achieve an improvement of 1.34\% and 0.58\% in validation accuracy on MNIST and Fashion-MNIST, respectively. On MNIST, \texttt{PRKAN-conv\&pool} exhibits the lowest validation accuracy and F1 score, while its training time is the second longest. On Fashion-MNIST, \texttt{PRKAN-dim-sum} shows the lowest validation accuracy and F1 score, but its training time is the second fastest, surpassed only by MLPs. Overall, batch normalization provides the most significant improvement to \texttt{PRKAN-attn}, enhancing both validation accuracy and F1 scores, whereas other PRKAN variants show no noticeable benefit from its application.

With layer normalization, although MLPs maintain the highest performance in terms of F1 and accuracy values, \texttt{PRKAN-attn} models deliver highly competitive results, surpassing other PRKAN variants and trailing only behind MLPs in accuracy and F1 scores. In more detail, on MNIST, \texttt{PRKAN-attn} achieves a validation accuracy of 97.46\%, trailing MLP by a margin of less than 0.26\%. A similar trend is observed on Fashion-MNIST, where \texttt{PRKAN-attn} is just 0.14\% behind MLP in validation accuracy. However, the average training time of \texttt{PRKAN-attn} models is approximately 10\% longer compared to MLPs. Moreover, they exhibit a drawback in terms of excessive FLOPs compared to MLPs, as illustrated in \Cref{fig:flops_vs_val_acc}.

Without data normalization, \texttt{PRKAN-attn} models exhibit unstable and significantly lower performance compared to other models, surpassing only \texttt{PRKAN-dim-sum}, which appears flawed with a validation accuracy below 12\%. MLPs and \texttt{PRKAN-conv} models dominate the top performance, with MLPs performing best on MNIST, while \texttt{PRKAN-conv} excels on Fashion-MNIST. Besides, MLPs have the shortest training time thanks to their simple architecture, followed by \texttt{PRKAN-dim-sum} and \texttt{PRKAN-fwv}.

Despite its fast training time and minimal use of FLOPs, \texttt{PRKAN-dim-sum} models exhibit a worse performance. This suggests that dimension summation after feature extraction may lead to the loss of important features. Other PRKAN variants, including \texttt{PRKAN-conv}, \texttt{PRKAN-conv\&pool}, and \texttt{PRKAN-fwv}, outperform \texttt{PRKAN-dim-sum}, but they do not show significant advantages over MLPs. Notably, \texttt{PRKAN-conv} and \texttt{PRKAN-conv\&pool} have the longest training times but do not achieve the highest metric values, emphasizing that the use of convolution layers to extract data features after applying functions used in KAN layers may not be an efficient method. 

\begin{figure*}[htbp]
  \centering
\includegraphics[scale=0.6]{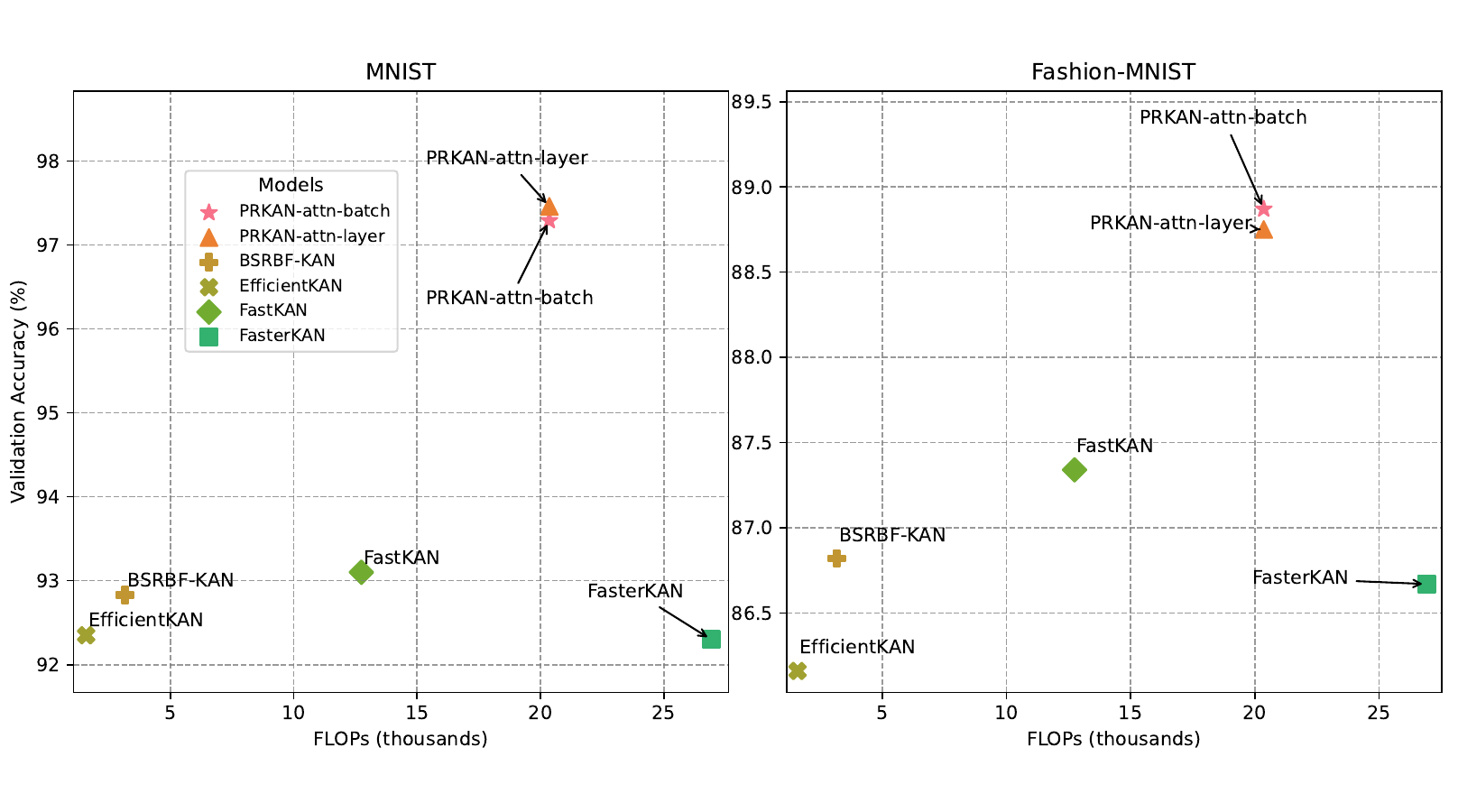}
  \centering
  \caption{\texttt{PRKAN-attn} models with batch and layer normalization are compared to other existing KANs in terms of FLOPs and validation accuracy.}
\label{fig:prkans_vs_others_flops_acc}
\end{figure*}

\subsection{PRKANs vs. Other Existing KANs}
\begin{table*}[ht]
	\caption{The comparison between PRKANs and other KANs.}
	\centering
	\begin{tabular}{p{1.5cm}p{2.5cm}p{1.5cm}p{1.8cm}p{1.8cm}p{1.8cm}p{1.4cm}}
            \hline
		\textbf{Dataset} &  \textbf{Model}  & \textbf{Norm.} & \textbf{Train. Acc.} & \textbf{Val. Acc.} & \textbf{F1} & \textbf{Time (sec)} \\
    \hline 
    \multirow{6}{1.5cm}{\textbf{MNIST}} 
    & PRKAN-attn & batch & 98.97 ± 0.33 & 97.29 ± 0.10 & 97.25 ± 0.10 & 179.35 \\
    & PRKAN-attn & layer & \textbf{99.81 ± 0.09} & \textbf{97.46 ± 0.06} & \textbf{97.42 ± 0.06} & 178.02 \\
    & BSRBF-KAN & layer & 95.33 ± 0.14 & 92.83 ± 0.19 & 92.68 ± 0.19 & 209.73 \\
    & EfficientKAN & -- & 93.33 ± 0.05 & 92.35 ± 0.12 & 92.24 ± 0.12 & 180.79 \\
    & FastKAN & layer & 95.00 ± 0.11 & 93.10 ± 0.22 & 92.97 ± 0.24 & 164.64 \\
    & FasterKAN & layer & 92.82 ± 0.06 & 92.30 ± 0.08 & 92.17 ± 0.09 & \textbf{155.30} \\
            \hline
            \hline
    \multirow{6}{1.5cm}{\textbf{Fashion-MNIST}} 
    & PRKAN-attn & batch & \textbf{94.10 ± 0.17} & \textbf{88.87 ± 0.06} & \textbf{88.82 ± 0.06} & 259.18 \\ 
    & PRKAN-attn & layer & 93.30 ± 0.20 & 88.82 ± 0.09 & 88.75 ± 0.10 & 250.62 \\
    & BSRBF-KAN & layer & 92.89 ± 0.07 & 86.82 ± 0.08 & 86.77 ± 0.08 & 295.71 \\
    & EfficientKAN & -- & 89.00 ± 0.08 & 86.16 ± 0.12 & 86.07 ± 0.12 & 254.25 \\
    & FastKAN & layer & 91.59 ± 0.07 & 87.34 ± 0.05 & 87.28 ± 0.04 & 229.21 \\
    & FasterKAN & layer & 89.16 ± 0.09 & 86.67 ± 0.12 & 86.57 ± 0.11 & \textbf{217.42} \\
    
            \hline
             \multicolumn{7}{l}{Norm. = Data Normalization, Train. Acc = Training Accuracy, Val. Acc. = Validation Accuracy }  \\
             \multicolumn{7}{l}{\texttt{attn} = attention mechanism}  \\
             \hline
	\end{tabular}
	\label{tab:prkans_vs_other_kans}
\end{table*}

To address the issue of unfair parameter allocation in existing KANs compared to MLPs, we modify the network structure of these models to ensure a comparable parameter count with both MLPs and PRKANs, as shown in \Cref{tab:model_params_flops}. For the comparison, we use the results of \texttt{PRKAN-attn} models with RFBs and MLPs, both using layer normalization, as presented in \Cref{tab:prkans_vs_mlps}. All other hyperparameters remain consistent with those detailed in \Cref{sec:training_conf} during the training of all KAN models.

EfficientKAN does not use any data normalization, while all other KANs apply layer normalization to the input. It’s important to note that this is the default configuration for these KANs. \Cref{tab:prkans_vs_other_kans} clearly illustrates that existing KANs cannot match the performance of PRKANs and MLPs when configured with a comparable number of parameters. The performance gap between PRKANs and other KANs is substantial, ranging from approximately 2.7\% to 5.2\% in validation accuracy, with PRKANs outperforming the others. Compared to PRKANs, FastKANs show lower performance, followed by BSRBF-KANs, which have the longest training time. FasterKANs, while offering the quickest training time, do not exhibit notable performance. Without layer normalization, EfficientKANs perform the worst overall and is unable to compete well with other KANs. 

The correlation of FLOPs and validation accuracy values is presented in \Cref{fig:prkans_vs_others_flops_acc} illustrates the trend that higher FLOPs generally correspond to higher accuracy. In this context, \texttt{PRKAN-attn} models achieve the best accuracy while also having the highest FLOPs. However, FasterKAN deviates from this trend, as it consumes significantly more FLOPs than \texttt{PRKAN-attn} but delivers much lower accuracy. In summary, PRKANs exhibit the most significant advantages over other KANs when configured with an equivalent parameter count.

\subsection{Ablation Study}


\subsubsection{Activation Functions}

\begin{table*}[ht]
\caption{Activation functions with formulas and key features.}
\centering
\begin{tabular}{p{7cm}p{7cm}}
\hline
\textbf{Activation Function} & \textbf{Formula} \\
\hline
\textbf{ELU (Exponential Linear Unit)} \newline
Reduces bias shift, offers smooth transition, and is customizable with $\alpha$.  
& $f(x) = \begin{cases} 
x & \text{if } x > 0 \\
\alpha(e^x - 1) & \text{if } x \leq 0
\end{cases}$ \\
\hline
\textbf{GELU (Gaussian Error Linear Unit)} \newline
Combines smoothness and non-linearity, is probabilistic, and excels in NLP tasks.  
& $f(x) = 0.5x \left(1 + \tanh\left(\sqrt{\frac{2}{\pi}}(x + 0.044715x^3)\right)\right)$ \\
\hline
\textbf{Leaky ReLU (Leaky Rectified Linear Unit)} \newline
Allows gradients for $x < 0$, avoids dead neurons, and is customizable with $\alpha$.  
& $f(x) = \begin{cases} 
x & \text{if } x > 0 \\
\alpha x & \text{if } x \leq 0
\end{cases}$ \\
\hline
\textbf{ReLU (Rectified Linear Unit)} \newline
Efficient, sparse activations, with the risk of dead neurons.  
& $f(x) = \max(0, x)$ \\
\hline
\textbf{SELU (Scaled Exponential Linear Unit)} \newline
Self-normalizing, scaled output, and requires specific initialization.  
& $f(x) = \lambda \begin{cases} 
x & \text{if } x > 0 \\
\alpha(e^x - 1) & \text{if } x \leq 0
\end{cases}$ \\
\hline
\textbf{Sigmoid} \newline
Smooth, bounded, maps input to (0, 1), and is commonly used in binary classification.  
& $f(x) = \frac{1}{1 + e^{-x}}$ \\
\hline
\textbf{SiLU (Sigmoid Linear Unit)} \newline
Smooth, self-gating, and enhances gradient flow.  
& $f(x) = \frac{x}{1 + e^{-x}}$ \\
\hline
\textbf{Softplus (Smooth ReLU Approximation)} \newline
Differentiable, smooth approximation of ReLU, and no hard zero cutoff.  
& $f(x) = \log(1 + e^x)$ \\
\hline
\end{tabular}
\label{tab:activation_features}
\end{table*}

We choose several activation functions listed in \Cref{tab:activation_features} and experiment with them during the training of PRKAN models. In KANs, activation functions primarily perform linear transformations after data features are extracted by outer functions (e.g., B-splines, RBFs). We also investigate whether other activation functions can improve the model's performance in terms of training time and validation accuracy. For the configuration, all \texttt{PRKAN-attn} models use layer normalization, RBFs, and other hyperparameters as specified in \Cref{sec:training_conf}. Note that each model is trained in a single independent run.

The results presented in \Cref{tab:act_functions} highlight the suitability of SiLU for our experiments, as it achieved the best validation accuracy and F1 score on the MNIST dataset while delivering competitive performance on Fashion-MNIST. Moreover, activation functions such as ELU, GELU, and SELU demonstrated strong performance, with GELU achieving the highest results on Fashion-MNIST. The time differences between these activation functions are minimal since all are well-optimized for use with PyTorch and GPU devices. ELU exhibited strong convergence capabilities but required the longest training times on both datasets. On the other hand, Sigmoid and Softplus are not recommended due to their poorest overall performance. 

\begin{table*}[ht]

    \caption{The performance of \texttt{PRKAN-attn} models (layer normalization, RBFs) by activation functions. Each model, with its corresponding activation function, was trained in a single run.}
    \centering
    \begin{tabular}{p{1.5cm}p{2cm}p{1.8cm}p{1.8cm}p{1.8cm}p{2.5cm}}
        \hline
        \textbf{Dataset} & \textbf{Act. Func.} & \textbf{Train. Acc.} & \textbf{Val. Acc.} & \textbf{F1} & \textbf{Time (sec)} \\
        \hline
        \multirow{8}{1.5cm}{\textbf{MNIST}} 
        & ELU & \textbf{99.99} & 97.39 & 97.35 & 182.46 \\
        & GELU & 99.77 & 97.51 & 97.47 & 181.25 \\
        & Leaky ReLU & 99.86 & 97.21 & 97.17 & 180.32 \\
        & ReLU & 98.19 & 97.24 & 97.20 & 176.63 \\
        & SELU & 99.98 & 97.44 & 97.40 & 178.00 \\
        & Sigmoid & 94.98 & 94.02 & 93.94 & \textbf{175.02} \\
        & SiLU & 99.71 & \textbf{97.59} & \textbf{97.55} & 180.14 \\
        & Softplus & 94.09 & 93.48 & 93.35 & 180.10 \\
        \hline
        \hline
        \multirow{8}{1.5cm}{\textbf{Fashion-MNIST}} 
        & ELU & \textbf{94.87} & 89.02 & 89.02 & 252.71 \\
        & GELU & 94.34 & \textbf{89.19} & 89.12 & 251.14 \\
        & Leaky ReLU & 92.84 & 88.88 & 88.76 & 252.21 \\
        & ReLU & 92.16 & 88.34 & 88.32 & 251.70 \\
        & SELU & 94.86 & 89.18 & \textbf{89.13} & 262.48 \\
        & Sigmoid & 87.59 & 85.60 & 85.48 & 253.75 \\
        & SiLU & 93.73 & 88.97 & 88.93 & 256.30 \\
        & Softplus & 87.41 & 85.79 & 85.67 & \textbf{250.87} \\
        \hline
        \multicolumn{6}{l}{Act. Func. = Activation Function, Train. Acc. = Training Accuracy, Val. Acc. = Validation Accuracy} \\
        \hline
    \end{tabular}
    \label{tab:act_functions}
\end{table*}

\subsubsection{B-spline vs. GRBF}
Though many basis functions are used in KANs to fit data, we only apply the code implementations of B-splines\footnote{\url{https://github.com/Blealtan/efficient-kan}} and GRBFs\footnote{\url{https://github.com/ZiyaoLi/fast-kan}} because they allow extracting data in the shape of \( (B, D, G + k) \) from a given input \( (B, D) \) without requiring additional parameters, making them suitable for parameter reduction methods. In contrast, other functions, such as wavelets in Wav-KAN, involve translation and scaling~\cite{bozorgasl2024wav}, resulting in more parameters. In this experiment, we use \texttt{PRKAN-attn} models with layer normalization, SiLU, and other hyperparameters as specified in \Cref{sec:training_conf}.

\begin{table*}[ht]
	\caption{The comparison of B-splines and RBFs in \texttt{PRKAN-attn} models.}
	\centering
	\begin{tabular}{p{1.5cm}p{1.5cm}p{1.5cm}p{1.8cm}p{1.8cm}p{1.8cm}p{2.5cm}}
		\hline
		\textbf{Dataset} & \textbf{Function} & \textbf{Norm.} & \textbf{Train. Acc.} & \textbf{Val. Acc.} & \textbf{F1} & \textbf{Time (sec)} \\
		\hline
		\multirow{4}{1.5cm}{\textbf{MNIST}} 
         & B-spline & batch & 99.02 ± 0.14 & 97.15 ± 0.05 & 97.11 ± 0.05 & 204.7 \\
        & B-spline & layer & 99.28 ± 0.12 & \textbf{97.54 ± 0.08} & \textbf{97.51 ± 0.08} & 205.3 \\
        & RBF & batch & 98.97 ± 0.33 & 97.29 ± 0.10 & 97.25 ± 0.10 & 179.35 \\
        & RBF & layer & \textbf{99.81 ± 0.09} & 97.46 ± 0.06 & 97.42 ± 0.06 & \textbf{178.02} \\
		\hline
		\hline
		\multirow{4}{1.5cm}{\textbf{Fashion-MNIST}} 
        & B-spline & batch & 92.05 ± 0.17 & 88.67 ± 0.12 & 88.60 ± 0.11 & 291.33 \\
        & B-spline & layer & 92.33 ± 0.17 & 88.60 ± 0.15 & 88.54 ± 0.15 & 284.51 \\	
        & RBF & batch & \textbf{94.10 ± 0.17} & \textbf{88.87 ± 0.06} & \textbf{88.82 ± 0.06} & 259.18 \\ 
        & RBF & layer & 93.30 ± 0.20 & 88.82 ± 0.09 & 88.75 ± 0.10 & \textbf{250.62} \\
		\hline
		\multicolumn{7}{l}{Norm. = Data Normalization, Train. Acc = Training Accuracy, Val. Acc. = Validation Accuracy}  \\
		\hline
	\end{tabular}
	\label{tab:bs_vs_rbf}
\end{table*}

Based on the experimental results in \Cref{tab:bs_vs_rbf}, we found that GRBFs are 11\% to 13\% faster than B-splines. This aligns with the findings of some works \cite{li2024kolmogorov,abueidda2024deepokan}, highlighting the time-efficiency advantages of GRBFs. When comparing the same dataset and data normalization, GRBFs generally outperform B-splines, except when layer normalization is applied on MNIST. Overall, GRBFs offer more advantages than B-splines when applied to KANs.

\subsubsection{Positions of Data Normalization in PRKANs}
\label{norm_positions}
In this section, we examine the impact of different positions and types of data normalization on the performance of PRKANs within parameter-reduced methods. To maintain efficiency and limit complexity, data normalization is applied only once per PRKAN layer. We evaluate two commonly used normalization techniques: batch normalization (BN) and layer normalization (LN). While our intuition indicates that the absence of data normalization may not improve model performance, we also train models without normalization to assess its impact.

\begin{table*}[ht]
	\caption{Positions for applying data normalization in PRKANs using parameter-reduced methods.}
	\centering
	\begin{tabular}{p{3cm}p{3.5cm}p{4cm}}
		\hline
		\textbf{Method} & \textbf{Position 1 (@1)} & \textbf{Position 2 (@2)} \\
		\hline
        PRKAN-attn & After "Input (B, D)" & After "Summation (B, D)" \\
        PRKAN-conv & After "Input (B, D)" & After "Squeeze (B, D)" \\
        PRKAN-conv\&pool & After "Input (B, D)" & After "Reshape (B, D)" \\
        PRKAN-dim-sum & After "Input (B, D)" & After "Summation (B, D)" \\
        PRKAN-fwv & After "Input (B, D)" & After "Mul (B, D)" \\
        MLP-base & After "Input (B, D)" & - \\
		\hline
		\multicolumn{3}{l}{\texttt{attn} = attention mechanism, \texttt{conv} = convolutional layers} \\
        \multicolumn{3}{l}{\texttt{conv\&pool} = convolutional \& pooling layers, \texttt{dim-sum} = dimension summation}  \\
        \multicolumn{3}{l}{\texttt{fwv} = feature weight vectors, \texttt{base} = MLP layers}  \\
       
		\hline
	\end{tabular}
	\label{tab:norm_positions}
\end{table*}

\begin{figure*}[htbp]
  \centering
\includegraphics[scale=0.65]{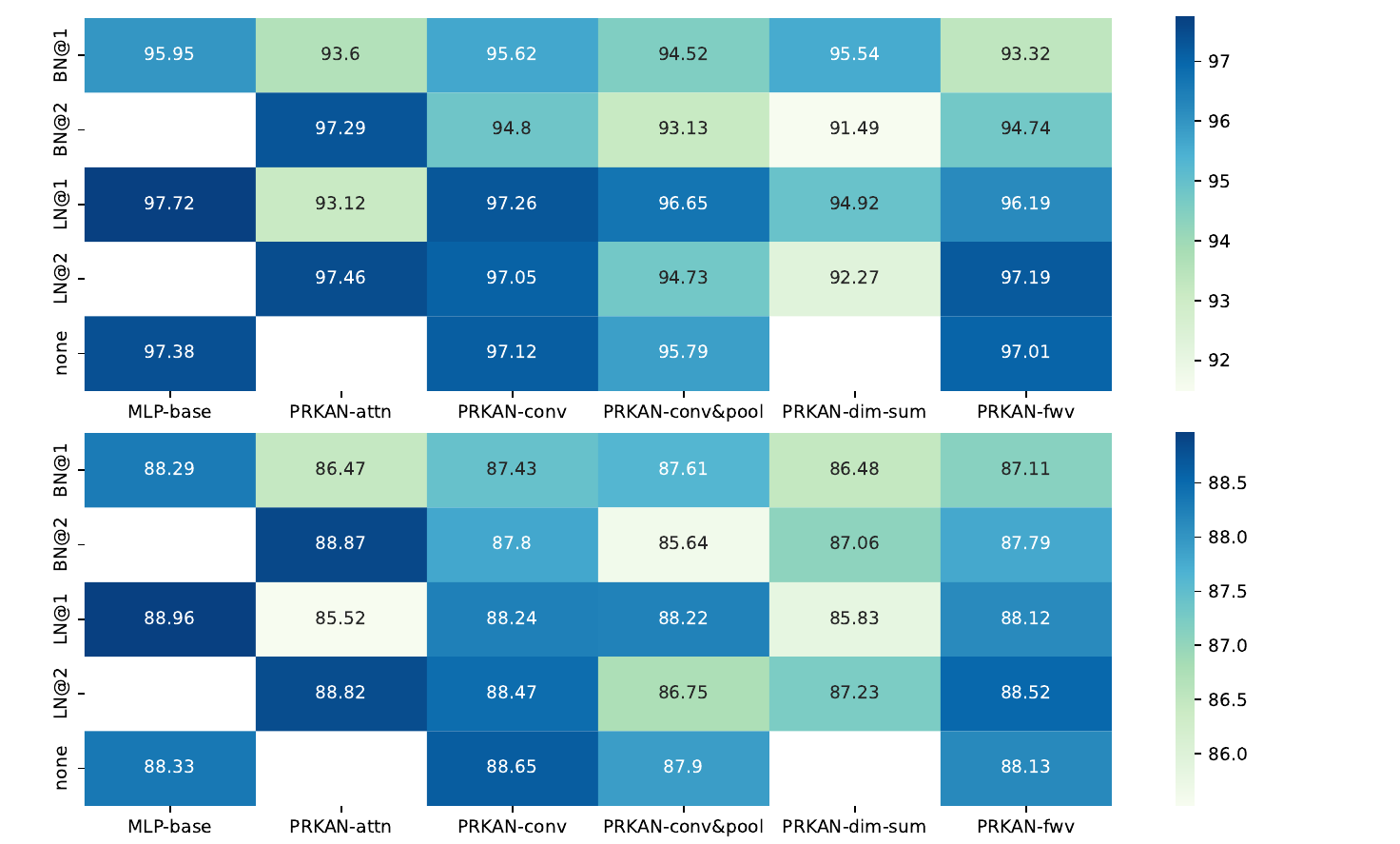}
  \centering
  \caption{Validation accuracy values for different models, categorized by data normalization methods and their positions. BN@1 and BN@2 denote batch normalization applied at positions 1 and 2, respectively, while LN@1 and LN@2 represent layer normalization at the same positions. Missing entries indicate models where validation accuracy is significantly lower, making them less relevant for comparison.}
\label{fig:heatmap_norm_pos_acc}
\end{figure*}

As shown in \Cref{fig:prkan_data_norm} and \Cref{data_norm}, data normalization is applied wherever the data tensor has the shape $(B, D)$, including the input. Several positions for incorporating normalization in parameter-reduced PRKAN methods are proposed, as experimented in \Cref{tab:norm_positions} and \Cref{appendix_data_norm_pos}. \texttt{MLP-base} models only apply data normalization after the input, while other models can apply it both after the input and at other positions with the shape $(B, D)$.

\Cref{fig:heatmap_norm_pos_acc} shows the validation accuracy of PRKAN models evaluated across different data normalization positions. Layer normalization is the most effective, consistently enhancing accuracy across various configurations. However, the specific data impact of normalization varies depending on data normalization positions and the parameter-reduction methods integrated into the PRKAN models. For example, \texttt{PRKAN-attn} achieve their best performance with layer normalization and batch normalization at position 2 but underperform in the absence of normalization. MLPs exhibit optimal accuracy with layer normalization applied at position 1. On the other hand, \texttt{PRKAN-conv} models prefer layer normalization at position 1 on MNIST but perform better without normalization on Fashion-MNIST. Similarly, \texttt{PRKAN-conv\&layer} models perform best with layer normalization at position 1. \texttt{PRKAN-dim-sum} models show no strong preference but achieve their highest accuracy with batch normalization at position 1 on MNIST and layer normalization at position 2 on Fashion-MNIST. Finally, \texttt{PRKAN-fwv} models perform optimally with layer normalization at position 2.

\section{Limitation}

The first limitation arises from the use of a shallow network structure (784, 64, 10) to deploy PRKANs with several parameter reduction methods. Although PRKANs perform competitively compared to MLPs in this shallow configuration, it is unclear whether PRKANs would outperform MLPs in deeper architectures. More research is needed to evaluate the scalability and efficiency of PRKANs in more complex models. Secondly, PRKANs have been tested on two popular but relatively simple datasets, MNIST and Fashion-MNIST. Although these data sets serve as effective benchmarks, they may not fully represent the challenges posed by more complex, high-dimensional data sets. Consequently, the generalizability of PRKANs to other datasets with more intricate patterns or dependencies remains uncertain.

The third limitation arises from the selection of methods for parameter reduction. Although several well-known methods, such as dimension summation, attention mechanisms, and convolutional layers, were examined, the study did not cover all possible ones. Methods such as tensor decomposition~\cite{ji2019survey,kolda2009tensor}, matrix factorization~\cite{sainath2013low}, and advanced pruning~\cite{cheng2024survey,vadera2022methods} could offer additional or complementary benefits. To address these limitations, future efforts could involve testing PRKANs in deeper network architectures, applying them to a wider variety of datasets, and exploring other parameter reduction strategies.

\section{Conclusion}

In this paper, we introduce PRKAN, a novel KAN that integrates various methods such as attention mechanisms, dimension summation, feature weight vectors, and convolutional and/or pooling layers to reduce the number of parameters in its layers, making it comparable to MLPs with a similar parameter count while maintaining the same network structure. In our experiments, we compare PRKANs with MLPs and other KANs using the MNIST and Fashion-MNIST datasets. With approximately the same number of parameters, PRKANs significantly outperform existing KANs. A variant of PRKAN incorporating an attention mechanism demonstrates competitive validation accuracy compared to MLPs. However, PRKANs face challenges compared to MLPs, including high FLOPs and increased training time, which can affect their overall efficiency. In addition, we present several ablation studies that compare outer functions, the effects of different activation functions, and data normalization strategies in parameter-reduced structures to achieve optimal results. We found that GRBFs and layer normalization generally provide more benefits when applied to PRKANs. 

Through this paper, we aim to take the first step toward exploring parameter reduction in Kolmogorov-Arnold Networks (KANs) by introducing recommended methods to simplify and optimize the architecture. Our goal is to develop efficient KANs that maintain high performance while significantly reducing the number of parameters. We also expect this work to lay the foundation for future research into more lightweight, high-performance KANs that can be applied across various domains, including image processing, natural language processing, and beyond. In the future, we will continue to investigate attention mechanisms in KANs, explore additional methods to reduce the number of parameters in KAN layers, such as tensor decomposition, and address the current limitations of PRKANs.

\section*{Acknowledgments}
We also acknowledge the support of (1) the Foundation for Science and Technology Development of Dalat University and (2) FPT University, Danang for funding this research.

\bibliographystyle{unsrtnat}
\bibliography{references}  

\begin{thebibliography}{81}
\providecommand{\natexlab}[1]{#1}
\providecommand{\url}[1]{\texttt{#1}}
\expandafter\ifx\csname urlstyle\endcsname\relax
  \providecommand{\doi}[1]{doi: #1}\else
  \providecommand{\doi}{doi: \begingroup \urlstyle{rm}\Url}\fi

\bibitem[Liu et~al.(2024{\natexlab{a}})Liu, Wang, Vaidya, Ruehle, Halverson, Solja{\v{c}}i{\'c}, Hou, and Tegmark]{liu2024kan}
Ziming Liu, Yixuan Wang, Sachin Vaidya, Fabian Ruehle, James Halverson, Marin Solja{\v{c}}i{\'c}, Thomas~Y Hou, and Max Tegmark.
\newblock Kan: Kolmogorov-arnold networks.
\newblock \emph{arXiv preprint arXiv:2404.19756}, 2024{\natexlab{a}}.

\bibitem[Liu et~al.(2024{\natexlab{b}})Liu, Ma, Wang, Matusik, and Tegmark]{liu2024kan2.0}
Ziming Liu, Pingchuan Ma, Yixuan Wang, Wojciech Matusik, and Max Tegmark.
\newblock Kan 2.0: Kolmogorov-arnold networks meet science.
\newblock \emph{arXiv preprint arXiv:2408.10205}, 2024{\natexlab{b}}.

\bibitem[Sternfeld(2006)]{sternfeld2006hilbert}
Yaki Sternfeld.
\newblock Hilbert's 13th problem and dimension.
\newblock In \emph{Geometric Aspects of Functional Analysis: Israel Seminar (GAFA) 1987--88}, pages 1--49. Springer, 2006.

\bibitem[Kolmogorov(1957)]{kolmogorov1957representation}
Andrei~Nikolaevich Kolmogorov.
\newblock On the representation of continuous functions of many variables by superposition of continuous functions of one variable and addition.
\newblock In \emph{Doklady Akademii Nauk}, volume 114, pages 953--956. Russian Academy of Sciences, 1957.

\bibitem[Li(2024)]{li2024kolmogorov}
Ziyao Li.
\newblock Kolmogorov-arnold networks are radial basis function networks.
\newblock \emph{arXiv preprint arXiv:2405.06721}, 2024.

\bibitem[Delis(2024)]{athanasios2024}
Athanasios Delis.
\newblock Fasterkan.
\newblock \url{https://github.com/AthanasiosDelis/faster-kan/}, 2024.

\bibitem[Ta(2024)]{ta2024bsrbf}
Hoang-Thang Ta.
\newblock Bsrbf-kan: A combination of b-splines and radial basis functions in kolmogorov-arnold networks.
\newblock \emph{arXiv preprint arXiv:2406.11173}, 2024.

\bibitem[Abueidda et~al.(2024)Abueidda, Pantidis, and Mobasher]{abueidda2024deepokan}
Diab~W Abueidda, Panos Pantidis, and Mostafa~E Mobasher.
\newblock Deepokan: Deep operator network based on kolmogorov arnold networks for mechanics problems.
\newblock \emph{arXiv preprint arXiv:2405.19143}, 2024.

\bibitem[Bhattacharjee(2024)]{torchkan}
Subhransu~S. Bhattacharjee.
\newblock Torchkan: Simplified kan model with variations.
\newblock \url{https://github.com/1ssb/torchkan/}, 2024.

\bibitem[SS(2024)]{ss2024chebyshev}
Sidharth SS.
\newblock Chebyshev polynomial-based kolmogorov-arnold networks: An efficient architecture for nonlinear function approximation.
\newblock \emph{arXiv preprint arXiv:2405.07200}, 2024.

\bibitem[Xu et~al.(2024{\natexlab{a}})Xu, Chen, Li, Yang, Wang, Hu, and Ngai]{xu2024fourierkan}
Jinfeng Xu, Zheyu Chen, Jinze Li, Shuo Yang, Wei Wang, Xiping Hu, and Edith C-H Ngai.
\newblock Fourierkan-gcf: Fourier kolmogorov-arnold network--an effective and efficient feature transformation for graph collaborative filtering.
\newblock \emph{arXiv preprint arXiv:2406.01034}, 2024{\natexlab{a}}.

\bibitem[Bozorgasl and Chen(2024)]{bozorgasl2024wav}
Zavareh Bozorgasl and Hao Chen.
\newblock Wav-kan: Wavelet kolmogorov-arnold networks.
\newblock \emph{arXiv preprint arXiv:2405.12832}, 2024.

\bibitem[Teymoor~Seydi(2024)]{teymoor2024exploring}
Seyd Teymoor~Seydi.
\newblock Exploring the potential of polynomial basis functions in kolmogorov-arnold networks: A comparative study of different groups of polynomials.
\newblock \emph{arXiv e-prints}, pages arXiv--2406, 2024.

\bibitem[Yu et~al.(2024)Yu, Yu, and Wang]{yu2024kan}
Runpeng Yu, Weihao Yu, and Xinchao Wang.
\newblock Kan or mlp: A fairer comparison.
\newblock \emph{arXiv preprint arXiv:2407.16674}, 2024.

\bibitem[Zeng et~al.(2024)Zeng, Wang, Shen, and Wang]{zeng2024kanversusmlpirregular}
Chen Zeng, Jiahui Wang, Haoran Shen, and Qiao Wang.
\newblock Kan versus mlp on irregular or noisy functions, 2024.
\newblock URL \url{https://arxiv.org/abs/2408.07906}.

\bibitem[Ta et~al.(2024)Ta, Thai, Rahman, Sidorov, and Gelbukh]{ta2024fc}
Hoang-Thang Ta, Duy-Quy Thai, Abu Bakar~Siddiqur Rahman, Grigori Sidorov, and Alexander Gelbukh.
\newblock Fc-kan: Function combinations in kolmogorov-arnold networks.
\newblock \emph{arXiv preprint arXiv:2409.01763}, 2024.

\bibitem[Yang et~al.(2024)Yang, Zhang, Luo, Lu, and Shen]{yang2024activation}
Zhuoqin Yang, Jiansong Zhang, Xiaoling Luo, Zheng Lu, and Linlin Shen.
\newblock Activation space selectable kolmogorov-arnold networks.
\newblock \emph{arXiv preprint arXiv:2408.08338}, 2024.

\bibitem[Moradi et~al.(2024)Moradi, Panahi, Bollt, and Lai]{moradi2024kolmogorov}
Mohammadamin Moradi, Shirin Panahi, Erik Bollt, and Ying-Cheng Lai.
\newblock Kolmogorov-arnold network autoencoders.
\newblock \emph{arXiv preprint arXiv:2410.02077}, 2024.

\bibitem[Sohail(2024)]{sohail2024training}
Shairoz Sohail.
\newblock On training of kolmogorov-arnold networks.
\newblock \emph{arXiv preprint arXiv:2411.05296}, 2024.

\bibitem[Shuai and Li(2024)]{shuai2024physics}
Hang Shuai and Fangxing Li.
\newblock Physics-informed kolmogorov-arnold networks for power system dynamics.
\newblock \emph{arXiv preprint arXiv:2408.06650}, 2024.

\bibitem[Braun and Griebel(2009)]{braun2009constructive}
J{\"u}rgen Braun and Michael Griebel.
\newblock On a constructive proof of kolmogorov’s superposition theorem.
\newblock \emph{Constructive approximation}, 30:\penalty0 653--675, 2009.

\bibitem[Zhou et~al.(2022)Zhou, Zhu, Wang, Ma, Wen, Sun, and Jin]{zhou2022treedrnet}
Tian Zhou, Jianqing Zhu, Xue Wang, Ziqing Ma, Qingsong Wen, Liang Sun, and Rong Jin.
\newblock Treedrnet: a robust deep model for long term time series forecasting.
\newblock \emph{arXiv preprint arXiv:2206.12106}, 2022.

\bibitem[Leni et~al.(2013)Leni, Fougerolle, and Truchetet]{leni2013kolmogorov}
Pierre-Emmanuel Leni, Yohan~D Fougerolle, and Fr{\'e}d{\'e}ric Truchetet.
\newblock The kolmogorov spline network for image processing.
\newblock In \emph{Image Processing: Concepts, Methodologies, Tools, and Applications}, pages 54--78. IGI Global, 2013.

\bibitem[Lai and Shen(2021)]{lai2021kolmogorov}
Ming-Jun Lai and Zhaiming Shen.
\newblock The kolmogorov superposition theorem can break the curse of dimensionality when approximating high dimensional functions.
\newblock \emph{arXiv preprint arXiv:2112.09963}, 2021.

\bibitem[van Deventer et~al.(2022)van Deventer, van Rensburg, and Bosman]{van2022kasam}
Heinrich van Deventer, Pieter~Janse van Rensburg, and Anna Bosman.
\newblock Kasam: Spline additive models for function approximation.
\newblock \emph{arXiv preprint arXiv:2205.06376}, 2022.

\bibitem[Girosi and Poggio(1989)]{girosi1989representation}
Federico Girosi and Tomaso Poggio.
\newblock Representation properties of networks: Kolmogorov's theorem is irrelevant.
\newblock \emph{Neural Computation}, 1\penalty0 (4):\penalty0 465--469, 1989.

\bibitem[Vitushkin(1954)]{vitushkin1954hilbert}
AG~Vitushkin.
\newblock On hilbert’s thirteenth problem.
\newblock In \emph{Dokl. Akad. Nauk SSSR}, volume~95, pages 701--704, 1954.

\bibitem[Lin and Unbehauen(1993)]{lin1993realization}
Ji-Nan Lin and Rolf Unbehauen.
\newblock On the realization of a kolmogorov network.
\newblock \emph{Neural Computation}, 5\penalty0 (1):\penalty0 18--20, 1993.

\bibitem[Kůrková(1991)]{kuurkova1991kolmogorov}
Věra Kůrková.
\newblock Kolmogorov's theorem is relevant.
\newblock \emph{Neural computation}, 3\penalty0 (4):\penalty0 617--622, 1991.

\bibitem[Dhiman(2024)]{dhiman2024kan}
Vikas Dhiman.
\newblock Kan: Kolmogorov--arnold networks: A review.
\newblock \url{https://vikasdhiman.info/reviews/KAN_a_review.pdf}, 2024.

\bibitem[Hao et~al.(2024)Hao, Zhang, Li, and Zhou]{hao2024first}
Hao Hao, Xiaoqun Zhang, Bingdong Li, and Aimin Zhou.
\newblock A first look at kolmogorov-arnold networks in surrogate-assisted evolutionary algorithms.
\newblock \emph{arXiv preprint arXiv:2405.16494}, 2024.

\bibitem[Wang et~al.(2024)Wang, Sun, Bai, Anitescu, Eshaghi, Zhuang, Rabczuk, and Liu]{wang2024kolmogorov}
Yizheng Wang, Jia Sun, Jinshuai Bai, Cosmin Anitescu, Mohammad~Sadegh Eshaghi, Xiaoying Zhuang, Timon Rabczuk, and Yinghua Liu.
\newblock Kolmogorov arnold informed neural network: A physics-informed deep learning framework for solving pdes based on kolmogorov arnold networks.
\newblock \emph{arXiv preprint arXiv:2406.11045}, 2024.

\bibitem[Koenig et~al.(2024)Koenig, Kim, and Deng]{koenig2024kan}
Benjamin~C Koenig, Suyong Kim, and Sili Deng.
\newblock Kan-odes: Kolmogorov--arnold network ordinary differential equations for learning dynamical systems and hidden physics.
\newblock \emph{Computer Methods in Applied Mechanics and Engineering}, 432:\penalty0 117397, 2024.

\bibitem[Xu et~al.(2024{\natexlab{b}})Xu, Zhang, Kong, Huang, Yang, Srivastava, and Sun]{xu2024effective}
Anfeng Xu, Biqiao Zhang, Shuyu Kong, Yiteng Huang, Zhaojun Yang, Sangeeta Srivastava, and Ming Sun.
\newblock Effective integration of kan for keyword spotting.
\newblock \emph{arXiv preprint arXiv:2409.08605}, 2024{\natexlab{b}}.

\bibitem[Kundu et~al.(2024)Kundu, Sarkar, and Sadhu]{kundu2024kanqas}
Akash Kundu, Aritra Sarkar, and Abhishek Sadhu.
\newblock Kanqas: Kolmogorov-arnold network for quantum architecture search.
\newblock \emph{EPJ Quantum Technology}, 11\penalty0 (1):\penalty0 76, 2024.

\bibitem[Wakaura and Suksmono(2024)]{wakaura2024variational}
H~Wakaura and AB~Suksmono.
\newblock Variational quantum kolmogorov-arnold network.
\newblock 2024.

\bibitem[Troy(2024)]{troy2024sparks}
William Troy.
\newblock Sparks of quantum advantage and rapid retraining in machine learning.
\newblock \emph{arXiv preprint arXiv:2407.16020}, 2024.

\bibitem[Knottenbelt et~al.(2024)Knottenbelt, Gao, Wray, Zhang, Liu, and Crispin-Ortuzar]{knottenbelt2024coxkan}
William Knottenbelt, Zeyu Gao, Rebecca Wray, Woody~Zhidong Zhang, Jiashuai Liu, and Mireia Crispin-Ortuzar.
\newblock Coxkan: Kolmogorov-arnold networks for interpretable, high-performance survival analysis.
\newblock \emph{arXiv preprint arXiv:2409.04290}, 2024.

\bibitem[Genet and Inzirillo(2024{\natexlab{a}})]{genet2024tkan}
Remi Genet and Hugo Inzirillo.
\newblock Tkan: Temporal kolmogorov-arnold networks.
\newblock \emph{arXiv preprint arXiv:2405.07344}, 2024{\natexlab{a}}.

\bibitem[Xu et~al.(2024{\natexlab{c}})Xu, Chen, and Wang]{xu2024kolmogorov}
Kunpeng Xu, Lifei Chen, and Shengrui Wang.
\newblock Kolmogorov-arnold networks for time series: Bridging predictive power and interpretability.
\newblock \emph{arXiv preprint arXiv:2406.02496}, 2024{\natexlab{c}}.

\bibitem[Vaca-Rubio et~al.(2024)Vaca-Rubio, Blanco, Pereira, and Caus]{vaca2024kolmogorov}
Cristian~J Vaca-Rubio, Luis Blanco, Roberto Pereira, and M{\`a}rius Caus.
\newblock Kolmogorov-arnold networks (kans) for time series analysis.
\newblock \emph{arXiv preprint arXiv:2405.08790}, 2024.

\bibitem[Genet and Inzirillo(2024{\natexlab{b}})]{genet2024temporal}
Remi Genet and Hugo Inzirillo.
\newblock A temporal kolmogorov-arnold transformer for time series forecasting.
\newblock \emph{arXiv preprint arXiv:2406.02486}, 2024{\natexlab{b}}.

\bibitem[Han et~al.(2024)Han, Zhang, Wu, Zhang, and Wu]{han2024kan4tsf}
Xiao Han, Xinfeng Zhang, Yiling Wu, Zhenduo Zhang, and Zhe Wu.
\newblock Kan4tsf: Are kan and kan-based models effective for time series forecasting?
\newblock \emph{arXiv preprint arXiv:2408.11306}, 2024.

\bibitem[Li et~al.(2024)Li, Liu, Li, Wang, Liu, and Yuan]{li2024u}
Chenxin Li, Xinyu Liu, Wuyang Li, Cheng Wang, Hengyu Liu, and Yixuan Yuan.
\newblock U-kan makes strong backbone for medical image segmentation and generation.
\newblock \emph{arXiv preprint arXiv:2406.02918}, 2024.

\bibitem[Cheon(2024)]{cheon2024demonstrating}
Minjong Cheon.
\newblock Demonstrating the efficacy of kolmogorov-arnold networks in vision tasks.
\newblock \emph{arXiv preprint arXiv:2406.14916}, 2024.

\bibitem[Ge et~al.(2024)Ge, Yu, Chen, Jia, Zhu, Zhou, Huang, Zhang, Zeng, Wang, et~al.]{ge2024tc}
Ruiquan Ge, Xiao Yu, Yifei Chen, Fan Jia, Shenghao Zhu, Guanyu Zhou, Yiyu Huang, Chenyan Zhang, Dong Zeng, Changmiao Wang, et~al.
\newblock Tc-kanrecon: High-quality and accelerated mri reconstruction via adaptive kan mechanisms and intelligent feature scaling.
\newblock \emph{arXiv preprint arXiv:2408.05705}, 2024.

\bibitem[Somvanshi et~al.(2024)Somvanshi, Javed, Islam, Pandit, and Das]{somvanshi2024survey}
Shriyank Somvanshi, Syed~Aaqib Javed, Md~Monzurul Islam, Diwas Pandit, and Subasish Das.
\newblock A survey on kolmogorov-arnold network.
\newblock \emph{arXiv preprint arXiv:2411.06078}, 2024.

\bibitem[De~Boor(1972)]{de1972calculating}
Carl De~Boor.
\newblock On calculating with b-splines.
\newblock \emph{Journal of Approximation theory}, 6\penalty0 (1):\penalty0 50--62, 1972.

\bibitem[Blealtan(2024)]{Blealtan2024}
Blealtan.
\newblock efficient-kan.
\newblock \url{https://github.com/Blealtan/efficient-kan}, 2024.

\bibitem[Seydi(2024)]{seydi2024unveiling}
Seyd~Teymoor Seydi.
\newblock Unveiling the power of wavelets: A wavelet-based kolmogorov-arnold network for hyperspectral image classification.
\newblock \emph{arXiv preprint arXiv:2406.07869}, 2024.

\bibitem[Aghaei(2024{\natexlab{a}})]{aghaei2024rkan}
Alireza~Afzal Aghaei.
\newblock rkan: Rational kolmogorov-arnold networks.
\newblock \emph{arXiv preprint arXiv:2406.14495}, 2024{\natexlab{a}}.

\bibitem[Aghaei(2024{\natexlab{b}})]{aghaei2024fkan}
Alireza~Afzal Aghaei.
\newblock fkan: Fractional kolmogorov-arnold networks with trainable jacobi basis functions.
\newblock \emph{arXiv preprint arXiv:2406.07456}, 2024{\natexlab{b}}.

\bibitem[Chen and Zhang(2024{\natexlab{a}})]{chen2024larctan}
Zhijie Chen and Xinglin Zhang.
\newblock Larctan-skan: Simple and efficient single-parameterized kolmogorov-arnold networks using learnable trigonometric function.
\newblock \emph{arXiv preprint arXiv:2410.19360}, 2024{\natexlab{a}}.

\bibitem[Chen and Zhang(2024{\natexlab{b}})]{chen2024lss}
Zhijie Chen and Xinglin Zhang.
\newblock Lss-skan: Efficient kolmogorov-arnold networks based on single-parameterized function.
\newblock \emph{arXiv preprint arXiv:2410.14951}, 2024{\natexlab{b}}.

\bibitem[Qiu et~al.(2024)Qiu, Zhu, Gong, Chen, and Ning]{qiu2024relu}
Qi~Qiu, Tao Zhu, Helin Gong, Liming Chen, and Huansheng Ning.
\newblock Relu-kan: New kolmogorov-arnold networks that only need matrix addition, dot multiplication, and relu.
\newblock \emph{arXiv preprint arXiv:2406.02075}, 2024.

\bibitem[Bresson et~al.(2024)Bresson, Nikolentzos, Panagopoulos, Chatzianastasis, Pang, and Vazirgiannis]{bresson2024kagnns}
Roman Bresson, Giannis Nikolentzos, George Panagopoulos, Michail Chatzianastasis, Jun Pang, and Michalis Vazirgiannis.
\newblock Kagnns: Kolmogorov-arnold networks meet graph learning.
\newblock \emph{arXiv preprint arXiv:2406.18380}, 2024.

\bibitem[De~Carlo et~al.(2024)De~Carlo, Mastropietro, and Anagnostopoulos]{de2024kolmogorov}
Gianluca De~Carlo, Andrea Mastropietro, and Aris Anagnostopoulos.
\newblock Kolmogorov-arnold graph neural networks.
\newblock \emph{arXiv preprint arXiv:2406.18354}, 2024.

\bibitem[Zhang and Zhang(2024)]{zhang2024graphkan}
Fan Zhang and Xin Zhang.
\newblock Graphkan: Enhancing feature extraction with graph kolmogorov arnold networks.
\newblock \emph{arXiv preprint arXiv:2406.13597}, 2024.

\bibitem[Kich et~al.(2024)Kich, Bottega, Steinmetz, Grando, Yorozu, and Ohya]{kich2024kolmogorov}
Victor~A Kich, Jair~A Bottega, Raul Steinmetz, Ricardo~B Grando, Ayano Yorozu, and Akihisa Ohya.
\newblock Kolmogorov-arnold networks for online reinforcement learning.
\newblock In \emph{2024 24th International Conference on Control, Automation and Systems (ICCAS)}, pages 958--963. IEEE, 2024.

\bibitem[Yang and Wang(2024)]{yang2024kolmogorov}
Xingyi Yang and Xinchao Wang.
\newblock Kolmogorov-arnold transformer.
\newblock \emph{arXiv preprint arXiv:2409.10594}, 2024.

\bibitem[Abd~Elaziz et~al.(2024)Abd~Elaziz, Fares, and Aseeri]{abd2024ckan}
Mohamed Abd~Elaziz, Ibrahim~Ahmed Fares, and Ahmad~O Aseeri.
\newblock Ckan: Convolutional kolmogorov--arnold networks model for intrusion detection in iot environment.
\newblock \emph{IEEE Access}, 2024.

\bibitem[Bodner et~al.(2024)Bodner, Tepsich, Spolski, and Pourteau]{bodner2024convolutional}
Alexander~Dylan Bodner, Antonio~Santiago Tepsich, Jack~Natan Spolski, and Santiago Pourteau.
\newblock Convolutional kolmogorov-arnold networks.
\newblock \emph{arXiv preprint arXiv:2406.13155}, 2024.

\bibitem[Danish and Grolinger(2025)]{danish2025kolmogorov}
Muhammad~Umair Danish and Katarina Grolinger.
\newblock Kolmogorov--arnold recurrent network for short term load forecasting across diverse consumers.
\newblock \emph{Energy Reports}, 13:\penalty0 713--727, 2025.

\bibitem[Le et~al.(2024)Le, Tran, Pham, Le, Vu, Nakashima, et~al.]{le2024exploring}
Tran Xuan~Hieu Le, Thi~Diem Tran, Hoai~Luan Pham, Vu~Trung~Duong Le, Tuan~Hai Vu, Yasuhiko Nakashima, et~al.
\newblock Exploring the limitations of kolmogorov-arnold networks in classification: Insights to software training and hardware implementation.
\newblock In \emph{2024 Twelfth International Symposium on Computing and Networking Workshops (CANDARW)}, pages 110--116. IEEE, 2024.

\bibitem[Ji et~al.(2019)Ji, Wang, Li, and Liu]{ji2019survey}
Yuwang Ji, Qiang Wang, Xuan Li, and Jie Liu.
\newblock A survey on tensor techniques and applications in machine learning.
\newblock \emph{IEEE Access}, 7:\penalty0 162950--162990, 2019.

\bibitem[Kolda and Bader(2009)]{kolda2009tensor}
Tamara~G Kolda and Brett~W Bader.
\newblock Tensor decompositions and applications.
\newblock \emph{SIAM review}, 51\penalty0 (3):\penalty0 455--500, 2009.

\bibitem[Sainath et~al.(2013)Sainath, Kingsbury, Sindhwani, Arisoy, and Ramabhadran]{sainath2013low}
Tara~N Sainath, Brian Kingsbury, Vikas Sindhwani, Ebru Arisoy, and Bhuvana Ramabhadran.
\newblock Low-rank matrix factorization for deep neural network training with high-dimensional output targets.
\newblock In \emph{2013 IEEE international conference on acoustics, speech and signal processing}, pages 6655--6659. IEEE, 2013.

\bibitem[Cheng et~al.(2024)Cheng, Zhang, and Shi]{cheng2024survey}
Hongrong Cheng, Miao Zhang, and Javen~Qinfeng Shi.
\newblock A survey on deep neural network pruning: Taxonomy, comparison, analysis, and recommendations.
\newblock \emph{IEEE Transactions on Pattern Analysis and Machine Intelligence}, 2024.

\bibitem[Vadera and Ameen(2022)]{vadera2022methods}
Sunil Vadera and Salem Ameen.
\newblock Methods for pruning deep neural networks.
\newblock \emph{IEEE Access}, 10:\penalty0 63280--63300, 2022.

\bibitem[Mou et~al.(2024)Mou, Xiao, Cao, Li, and Chen]{mou2024efficient}
Lanxin Mou, Xiongtao Xiao, Wenming Cao, Weikai Li, and Xiaofeng Chen.
\newblock Efficient and accurate capsule networks with b-spline-based activation functions.
\newblock In \emph{2024 International Conference on New Trends in Computational Intelligence (NTCI)}, pages 201--205. IEEE, 2024.

\bibitem[Pourkamali-Anaraki(2024)]{pourkamali2024kolmogorov}
Farhad Pourkamali-Anaraki.
\newblock Kolmogorov-arnold networks in low-data regimes: A comparative study with multilayer perceptrons.
\newblock \emph{arXiv preprint arXiv:2409.10463}, 2024.

\bibitem[Zinage et~al.(2024)Zinage, Mondal, and Sarkar]{zinage2024dkl}
Shrenik Zinage, Sudeepta Mondal, and Soumalya Sarkar.
\newblock Dkl-kan: Scalable deep kernel learning using kolmogorov-arnold networks.
\newblock \emph{arXiv preprint arXiv:2407.21176}, 2024.

\bibitem[Chernov(2020)]{chernov2020gaussian}
Andrei~Vladimirovich Chernov.
\newblock Gaussian functions combined with kolmogorov’s theorem as applied to approximation of functions of several variables.
\newblock \emph{Computational Mathematics and Mathematical Physics}, 60:\penalty0 766--782, 2020.

\bibitem[Schmidt-Hieber(2021)]{schmidt2021kolmogorov}
Johannes Schmidt-Hieber.
\newblock The kolmogorov--arnold representation theorem revisited.
\newblock \emph{Neural networks}, 137:\penalty0 119--126, 2021.

\bibitem[Barsky and Beatty(1983)]{barsky1983local}
Brian~A Barsky and John~C Beatty.
\newblock Local control of bias and tension in beta-splines.
\newblock In \emph{Proceedings of the 10th annual conference on Computer graphics and interactive techniques}, pages 193--218, 1983.

\bibitem[Liu et~al.(2019)Liu, Nassar, and Podg{\'o}rski]{liu2019splinets}
Xijia Liu, Hiba Nassar, and Krzysztof Podg{\'o}rski.
\newblock Splinets--efficient orthonormalization of the b-splines.
\newblock \emph{arXiv preprint arXiv:1910.07341}, 2019.

\bibitem[Liu et~al.(2021)Liu, Shao, and Hoffmann]{liu2021global}
Yichao Liu, Zongru Shao, and Nico Hoffmann.
\newblock Global attention mechanism: Retain information to enhance channel-spatial interactions.
\newblock \emph{arXiv preprint arXiv:2112.05561}, 2021.

\bibitem[Ioffe(2015)]{ioffe2015batch}
Sergey Ioffe.
\newblock Batch normalization: Accelerating deep network training by reducing internal covariate shift.
\newblock \emph{arXiv preprint arXiv:1502.03167}, 2015.

\bibitem[Ba(2016)]{ba2016layer}
Jimmy~Lei Ba.
\newblock Layer normalization.
\newblock \emph{arXiv preprint arXiv:1607.06450}, 2016.

\bibitem[Deng(2012)]{deng2012mnist}
Li~Deng.
\newblock The mnist database of handwritten digit images for machine learning research [best of the web].
\newblock \emph{IEEE signal processing magazine}, 29\penalty0 (6):\penalty0 141--142, 2012.

\bibitem[Xiao et~al.(2017)Xiao, Rasul, and Vollgraf]{xiao2017fashion}
Han Xiao, Kashif Rasul, and Roland Vollgraf.
\newblock Fashion-mnist: a novel image dataset for benchmarking machine learning algorithms.
\newblock \emph{arXiv preprint arXiv:1708.07747}, 2017.

\end{thebibliography}

\newpage
\appendix
\section{Performance by Data Normalization Positions}
\label{appendix_data_norm_pos}
\begin{table*}[ht]
	\caption{Performance of PRKAN models based on data normalization positions on MNIST. Each model was trained in a single run.}
	\centering
	\begin{tabular} 
    {p{2cm}p{1.5cm}p{1.5cm}p{2cm}p{2cm}p{2cm}p{2cm}}
		\hline
		\textbf{Method} & \textbf{Norm.} & \textbf{Position} & \textbf{Train. Acc.} & \textbf{Val. Acc.}  & \textbf{F1} & \textbf{Time (sec)}
        \\
        \hline
		\multirow{5}{2cm}{PRKAN-attn} 
        & batch & 1 & 97.06 ± 0.05 &  93.60 ± 0.42 & 93.53 ± 0.42 & 188.9 \\
        & batch & 2 & 98.97 ± 0.33 & 97.29 ± 0.10  & 97.25 ± 0.10 & 179.35 \\
        & layer & 1 & 93.57 ± 0.45 &  93.12 ± 0.45 & 93.02 ± 0.45 & 188.33 \\
        & layer & 2 & \textbf{99.81 ± 0.09} & \textbf{97.46 ± 0.06} & \textbf{97.42 ± 0.06} & 178.02 \\
        & none & - & 75.31 ± 5.52 & 75.36 ± 5.67 & 73.49 ± 6.36 & \textbf{176.45}
 \\
        \hline
        \multirow{5}{2cm}{PRKAN-conv} 
        & batch & 1 & 98.51 ± 0.58 & 95.62 ± 0.30 &  95.56 ± 0.30 &  199.73 \\
        & batch & 2 & 97.74 ± 0.46 & 94.80 ± 0.73  & 94.72 ± 0.75 & 187.05 \\
        & layer & 1 & \textbf{99.85 ± 0.09} & \textbf{97.26 ± 0.10} & \textbf{97.23 ± 0.10} & 193.07 \\
        & layer & 2 & \textbf{99.85 ± 0.07} &  97.05 ± 0.21  & 97.01 ± 0.22 & 184.83 \\
        & none & - &  99.66 ± 0.07 & 97.12 ± 0.10 & 97.08 ± 0.10 & \textbf{179.13} \\
        \hline
        \multirow{5}{2cm}{PRKAN-conv\&pool} 
        & batch & 1 & 97.22 ± 0.42 & 94.52 ± 0.44 & 94.45 ± 0.44 & 196.03 \\
        & batch & 2 & 97.91 ± 0.85 & 93.13 ± 0.30  & 93.07 ± 0.29 & 191.8 \\
        & layer & 1 & \textbf{99.47 ± 0.10}  & \textbf{96.65 ± 0.23} & \textbf{96.61 ± 0.24} & 196.5 \\
        & layer & 2 & 97.19 ± 0.32  & 94.73 ± 0.15  &  94.70 ± 0.15  &  187.05 \\
       & none & - & 98.85 ± 0.11 & 95.79 ± 0.07 &  95.75 ± 0.06 &  \textbf{183.78}  \\
        \hline
        \multirow{5}{2cm}{PRKAN-dim-sum} 
        & batch & 1 & \textbf{99.19 ± 0.15} &  \textbf{95.54 ± 0.06} & \textbf{95.50 ± 0.06} & 180.51 \\
        & batch & 2 & 98.60 ± 0.69 & 91.49 ± 0.11 & 91.42 ± 0.11  & 171.04 \\
        & layer & 1 & 95.71 ± 0.12 & 94.92 ± 0.13 & 94.83 ± 0.14 & 175.03 \\
        & layer & 2 & 99.05 ± 0.16  & 92.27 ± 0.13 & 92.18 ± 0.14 & 166.78 \\
        & none & - & 10.92 ± 0.17 &  11.36 ± 0.00 & 2.04 ± 0.00 &  \textbf{165.42} \\
        \hline
        \multirow{5}{2cm}{PRKAN-fwv} 
        & batch & 1 & 97.62 ± 0.85 &  93.32 ± 0.96 & 93.21 ± 0.97 & 179.7 \\
        & batch & 2 & 98.94 ± 0.62 & 94.74 ± 0.97 & 94.66 ± 0.99 &  171.98 \\
        & layer & 1 & \textbf{99.74 ± 0.15} & 96.19 ± 0.28 & 96.14 ± 0.29 & 182.13 \\
        & layer & 2 & 99.58 ± 0.12 & \textbf{97.19 ± 0.09} & \textbf{97.15 ± 0.09} & 168.92 \\
        & none & - & 99.46 ± 0.14 &  97.01 ± 0.13 & 96.97 ± 0.13 & \textbf{165.07} \\
        \hline
        \multirow{3}{2cm}{MLP-base} 
        & batch & 1 & 98.72 ± 0.23 & 95.95 ± 0.05 & 95.91 ± 0.05 & 165.84
 \\
        & layer & 1 & \textbf{99.84 ± 0.04} & \textbf{97.72 ± 0.05} & \textbf{97.69 ± 0.05} & 162.58 \\
        & none & - & 98.40 ± 0.04 &  97.38 ± 0.07 &  97.35 ± 0.07 & \textbf{156.68} \\
        
		\hline
		\multicolumn{7}{l}{\texttt{attn} = attention mechanism, \texttt{conv} = convolutional layers} \\
        \multicolumn{7}{l}{\texttt{conv\&pool} = convolutional \& pooling layers, \texttt{dim-sum} = dimension summation}  \\
        \multicolumn{7}{l}{\texttt{fwv} = feature weight vectors, \texttt{base} = MLP layers}  \\
        \hline
        
	\end{tabular}
	\label{tab:norm_pos_mnist}
\end{table*}

\begin{table*}[ht]
	\caption{Performance of PRKAN models based on data normalization positions on Fashion-MNIST. Each model was trained in a single run.}
	\centering
	\begin{tabular} 
    {p{2cm}p{1.5cm}p{1.5cm}p{2cm}p{2cm}p{2cm}p{2cm}}
		\hline
		\textbf{Method} & \textbf{Norm.} & \textbf{Position} & \textbf{Train. Acc.} & \textbf{Val. Acc.}  & \textbf{F1} & \textbf{Time (sec)}
        \\
        \hline
		\multirow{5}{2cm}{PRKAN-attn} 
        & batch & 1 &  90.26 ± 0.28 & 86.47 ± 0.29  & 86.29 ± 0.30 & 262.12 \\
        & batch & 2 & \textbf{94.10 ± 0.17} & \textbf{88.87 ± 0.06}  & \textbf{88.82 ± 0.06} & 259.18\\
        & layer & 1 &  87.86 ± 0.20 &  85.52 ± 0.17 & 85.39 ± 0.18 & 261.61 \\
        & layer & 2 & 93.30 ± 0.20 & 88.82 ± 0.09 & 88.75 ± 0.10 & 250.62 \\
        & none & - &  59.35 ± 3.82 & 58.96 ± 3.78 & 54.15 ± 4.75 & \textbf{249.33}
 \\
        \hline
        \multirow{5}{2cm}{PRKAN-conv} 
        & batch & 1 & 93.99 ± 1.27 & 87.43 ± 0.26 &   87.34 ± 0.27 &   275.25 \\
        & batch & 2 & 94.23 ± 0.66 & 87.80 ± 0.25  &  87.73 ± 0.26  & 264.08  \\
        & layer & 1 & \textbf{95.35 ± 0.61} & 88.24 ± 0.24 & 88.19 ± 0.25  & 273.85 \\
        & layer & 2 & 93.76 ± 0.10 & 88.47 ± 0.12 & 88.43 ± 0.12 & 260.52 \\
        & none & - &   92.65 ± 0.14 & \textbf{88.65 ± 0.06} & \textbf{88.58 ± 0.06} &  \textbf{257.22} \\
        \hline
        \multirow{5}{2cm}{PRKAN-conv\&pool} 
        & batch & 1 & 92.96 ± 1.18 & 87.61 ± 0.06 & 87.56 ± 0.06 & 278.91 \\
        & batch & 2 & 91.54 ± 0.67  & 85.64 ± 0.11  &  85.57 ± 0.10 & 268.72  \\
        & layer & 1 & \textbf{94.82 ± 0.42}  & \textbf{88.22 ± 0.07}  & \textbf{88.17 ± 0.08} & 278.12 \\
        & layer & 2 &  90.97 ± 0.15 & 86.75 ± 0.09 &  86.72 ± 0.08 & 263.88 \\
       & none & - &  94.08 ± 0.18 &  87.90 ± 0.08 &  87.87 ± 0.09 & \textbf{256.02}
  \\
        \hline
        \multirow{5}{2cm}{PRKAN-dim-sum} 
        & batch & 1 & \textbf{94.00 ± 0.42} &   86.48 ± 0.08 & 86.36 ± 0.09 & 245.92 \\
        & batch & 2 & 92.35 ± 0.09 & 87.06 ± 0.06 & 86.96 ± 0.06  & 244.03 \\
        & layer & 1 & 88.60 ± 0.08 & 85.83 ± 0.06 &  85.72 ± 0.06  & 239.54 \\
        & layer & 2 & 92.22 ± 0.05  & \textbf{87.23 ± 0.04} & \textbf{87.19 ± 0.05} & 238.64 \\
        & none & - & 9.86 ± 0.06 &  10.02 ± 0.01 & 1.82 ± 0.00 &  \textbf{236.59}  \\
        \hline
        \multirow{5}{2cm}{PRKAN-fwv} 
        & batch & 1 & 93.09 ± 0.93 &  87.11 ± 0.27 & 86.99 ± 0.29 & 252.36 \\
        & batch & 2 &  94.05 ± 0.75  &  87.79 ± 0.23 & 87.74 ± 0.23 & 245.86 \\
        & layer & 1 & \textbf{94.45 ± 0.46} & 88.12 ± 0.12 & 88.07 ± 0.11 & 255.33 \\
        & layer & 2 & 94.06 ± 0.27 & \textbf{88.52 ± 0.08} & \textbf{88.48 ± 0.08} & 241.93 \\
        & none & - & 92.82 ± 0.26 &   88.13 ± 0.23 & 88.07 ± 0.24 & \textbf{234.16} \\
        \hline
        \multirow{3}{2cm}{MLP-base} 
        & batch & 1 & \textbf{94.80 ± 0.06} &  88.29 ± 0.06 & 88.25 ± 0.06 & 233.71 \\
        & layer & 1 & 94.20 ± 0.09 & \textbf{88.96 ± 0.05} & \textbf{88.92 ± 0.05} & 226.79 \\
        & none & - & 91.30 ± 0.05 &  88.33 ± 0.03  &  88.25 ± 0.02 & \textbf{221.25} \\
		\hline
		\multicolumn{7}{l}{\texttt{attn} = attention mechanism, \texttt{conv} = convolutional layers} \\
        \multicolumn{7}{l}{\texttt{conv\&pool} = convolutional \& pooling layers, \texttt{dim-sum} = dimension summation}  \\
        \multicolumn{7}{l}{\texttt{fwv} = feature weight vectors, \texttt{base} = MLP layers}  \\
        \hline
	\end{tabular}
	\label{tab:norm_pos_fashion_mnist}
\end{table*}






\end{document}